\documentclass[runningheads]{llncs}
\usepackage{graphicx}
\usepackage{subfigure}
\usepackage[boxruled]{algorithm2e}
\usepackage{algorithmic}
\usepackage{bm}
\usepackage{amsfonts}
\usepackage{wrapfig}
\usepackage{float}
\begin{document}
%
\title{A novel joint points and silhouette-based method to estimate 3D human pose and shape}
%
%
\author{Zhongguo Li \and
Anders Heyden\and
Magnus Oskarsson}
\authorrunning{Z. Li et al.}
%
\institute{Lund University, Lund, Sweden \\
\email{zhongguo.li@math.lth.se, anders.heyden@math.lu.se, magnus.oskarsson@math.lth.se}}
\maketitle              
\begin{abstract}
This paper presents a novel method for 3D human pose and shape estimation from images with sparse views, using joint points and silhouettes, based on a parametric model. Firstly, the parametric model is fitted to the joint points estimated by deep learning-based human pose estimation. Then, we extract the correspondence between the parametric model of pose fitting and silhouettes on 2D and 3D space. A novel energy function based on the correspondence is built and minimized to fit parametric model to the silhouettes. Our approach uses sufficient shape information because the energy function of silhouettes is built from both 2D and 3D space. This also means that our method only needs images from sparse views, which balances data used and the required prior information. Results on synthetic data and real data demonstrate the competitive performance of our approach on pose and shape estimation of the human body.
\end{abstract}
\section{Introduction}
Estimation of 3D human body models from images is an important but challenging task in computer vision. In many practical fields, for instance, video games, VR/AR, E-commerce and biomedical research, 3D human body models are needed and play vital roles. However, the human body in real scenes naturally exhibits many challenging properties, such as non-rigid motion, clothes and occlusion. These factors make it difficult to accurately and efficiently estimate the 3D human body model from images, and many approaches have been proposed to obtain 3D human body models during the past decades. 

Time-of-flight cameras, and other types of hardware solutions, can provide depth information and have been one of the solutions to the reconstruction of 3D human bodies ~\cite{Izadi2011,Newcombe2015,Slavcheva2017,Yu_2018_CVPR,Xu2019}. More specifically, depth cameras are utilized to capture RGB images and the corresponding depth images of the scenes. The 3D meshes of each view can be computed from the RGB-D images and the complete 3D model can be estimated by fusing the 3D meshes of each view. The process of fusion is often implemented using the Iterated Closest Point (ICP) algorithm ~\cite{Izadi2011} or other similar improved algorithms, which often are computation-consuming. Since these methods only can handle rigid scenes well, research on dynamic scenes has been explored~\cite{Newcombe2015,Slavcheva2017,Yu_2018_CVPR,Xu2019}. However, compared to ordinary cameras, the cameras with depth sensors are still expensive and calibration of the depth camera can also be complicated. 

With the development of deep learning architectures, 3D human body models can be estimated by  optimization- \cite{Bogo_2016,alldieck_2018} or regression-based methods \cite{hmrKanazawa17,kolotouros_2019}. For methods based on optimization, prior information, for example, human poses and silhouettes can be estimated by deep neural networks. The 3D model can then be obtained by fitting the parametric human body model to the prior information. The regression-based methods use deep neural networks to directly estimate the parameters of the given parametric human body model from images, by training the deep neural networks\cite{hmrKanazawa17,varol18_2018,kolotouros_2019}. Both approaches have been explored extensively and have achieved good performance in 3D human body reconstruction. However, regression-based methods require a large amount of data to train the neural network. This often requires much work and it is difficult, and sometimes expensive, to generate the dataset. Compared to regression-based methods, the human pose estimation and semantic segmentation based on deep neural networks have been well developed and many pre-trained models can be utilized directly. This means that prior information in optimization-based methods can be more easily estimated through deep neural networks. For these reasons, our choice of method, proposed in this paper, is also optimization-based. 

In this paper, the goal is to estimate the 3D human body from images. Since this is a very complex problem and one single image can only provide limited prior information, a number of images taken from different view-points are used in our paper. The human pose estimation based on deep neural network ~\cite{xiao2018simple} is adopted to estimate the joint points of the human body in the multiple-view images. The Skinned multi-person linear model (SMPL) ~\cite{Loper_2015}, which is widely used in the methods based on optimization, is the parametric human body model also used in our paper. Then, an energy function is established based on the predicted joint points and the SMPL. By minimizing the energy function, we can achieve an estimated 3D human body model which has a pose consisted with the observed images. Afterwards, the silhouettes are exploited to improve the shape of the estimated human body model. Through building the correspondence edges between the estimated human body model and the given silhouettes from 2D and 3D space, the energy function for the silhouettes is constructed. The shape parameters of the human body model are obtained by optimizing the energy function. The final 3D human body model is generated by the estimated pose and shape parameters after pose fitting and shape fitting. The experiments on synthetic data and a public real dataset validate the performance of our method. 

In summary, the contribution of our method consists of two parts. Firstly, an improved energy function for silhouettes is constructed from 2D and 3D perspectives to estimate the parameters of shape. Secondly, a small number of images (four in our experiments) from different views are applied in our method, which balances the number of images and the prior information.

\section{Related Work}
In order to obtain 3D human body models from images, researchers have explored a large number of methods from hardware and software during the past decades. These work can be basically categorized according to whether a parametric human body model is adopted in the methods. For the approaches which do not depend on any parametric human body model, the 3D reconstruction of the human body is mainly implemented from RGB-D images captured by depth cameras. In contrast to the above methods, the approaches based on a parametric human body model often attempt to estimate the pose and shape from common RGB images. We call the two categories  \emph{parametric model-free} and \emph{parametric model based} methods, respectively.

Parametric model-free methods often reconstruct 3D human body models from RGB-D images, which means that these methods often require depth cameras. KinectFusion \cite{Izadi2011} was the typical work which used a Kinect depth camera to reconstruct the 3D meshes of an indoor scene with static objects. However, KinectFusion was mainly aiming at reconstructing rigid objects rather than dynamic scene like a moving human body. In order to tackle non-rigid reconstruction, DynamicFusion \cite{Newcombe2015}, VolumeDeform \cite{Innmann_2016}, KillingFusion \cite{Slavcheva2017} were proposed over the next several years. These methods can handle reconstruction of non-rigid and moving objects, but they typically only obtain good performance for partial body or small slow moving objects. Yu et al. proposed BodyFusion \cite{Yu_ICCV_2017} and DoubleFusion \cite{Yu_2018_CVPR} to reconstruct the whole 3D human body model for moving persons with high accuracy. One common thing in all of the above work is that they utilize one single Kinect to recover the 3D human body model. In order to improve the accuracy more, some methods based on multiple Kinects \cite{Ye_2012_ECCV,Dou_2016} were proposed to reconstruct 3D geometry, which was more complicated to set than single a Kinect. In addition, commercial depth \cite{Xu2019} cameras have also been used as a tool to reconstruct 3D models. The core idea of the parametric-model free methods is that they utilized depth cameras to capture RGB-D images and fused the meshes of each view to obtain the final 3D model. Although the work has achieved good performance for 3D reconstruction of human body, cameras with depth sensor are still inconvenient in many applications. 

Parametric model based methods often tackle the problem through fitting a parametric human body model to prior information of the given images. The parametric human body model is often trained by a dataset and is defined as a function of variables which can represent prior information like pose and shape. The parametric models such as SCAPE \cite{Anguelov_2005} and SMPL \cite{Loper_2015} have been used in many methods. Recently, an improved model called SMPL-X was proposed by considering the motion of face and hands \cite{Pavlakos_2019}. With the development of deep learning, some methods exploited deep neural networks to regress the parameters of the parametric model, hence we call them regression based methods. In \cite{hmrKanazawa17}, the pose and shape parameters of the SMPL model were estimated by training a deep encoder network. In \cite{varol18_2018}, the loss function of mesh was added to further finetune the mesh of the 3D model. In \cite{pavlakos2018}, the pose and shape parameters were separately trained in two pipelines to make the result better. Nikos et al. \cite{kolotouros_2019} used the output of deep neural network to initialize the SMPL model and then supervise the training process of deep neural networks through the SMPL model. In \cite{pavlakos2019}, texture was utilized to capitalizes on the appearance constancy of images from different viewpoints. Although these methods have achieved competitive results, collecting datasets for training is still cumbersome work and training the network is also time-consuming. Another way to solve the problem is to fit the parametric human body model  to prior information through optimizing an error function (optimization based methods). Early work \cite{LeonidNIPS2007} used SCAPE to estimate the articulated pose and non-rigid shape. In \cite{Guan_2009}, silhouettes and joint points were manually obtained and the SCAPE model was fitted to the priori clues to estimate the parameters of SCAPE. In \cite{Weiss_2011,Bogo_ICCV_2015}, RGB-D images were utilized to estimate the parameters of SCAPE model. Xu et al.\cite{Weipeng_2017} scanned a template as the parametric model and used it to fit the prior information through optimizing an energy function. Bogo et al \cite{Bogo_2016} proposed a method called SMPLify in which the joint points were predicted by human pose estimation based on a deep neural network, and then SMPL was fitted to the estimated joint points. Moreover, silhouettes \cite{alldieck_2018} and multiple images with different views \cite{Huang_2017,Li_2019} were introduced as prior information for the SMPL model. Overall, optimization based methods are often easier to implement, since it is unnecessary to create datasets and to do training.
\begin{figure}
	\centering
	\centering
	\includegraphics[width=\textwidth]{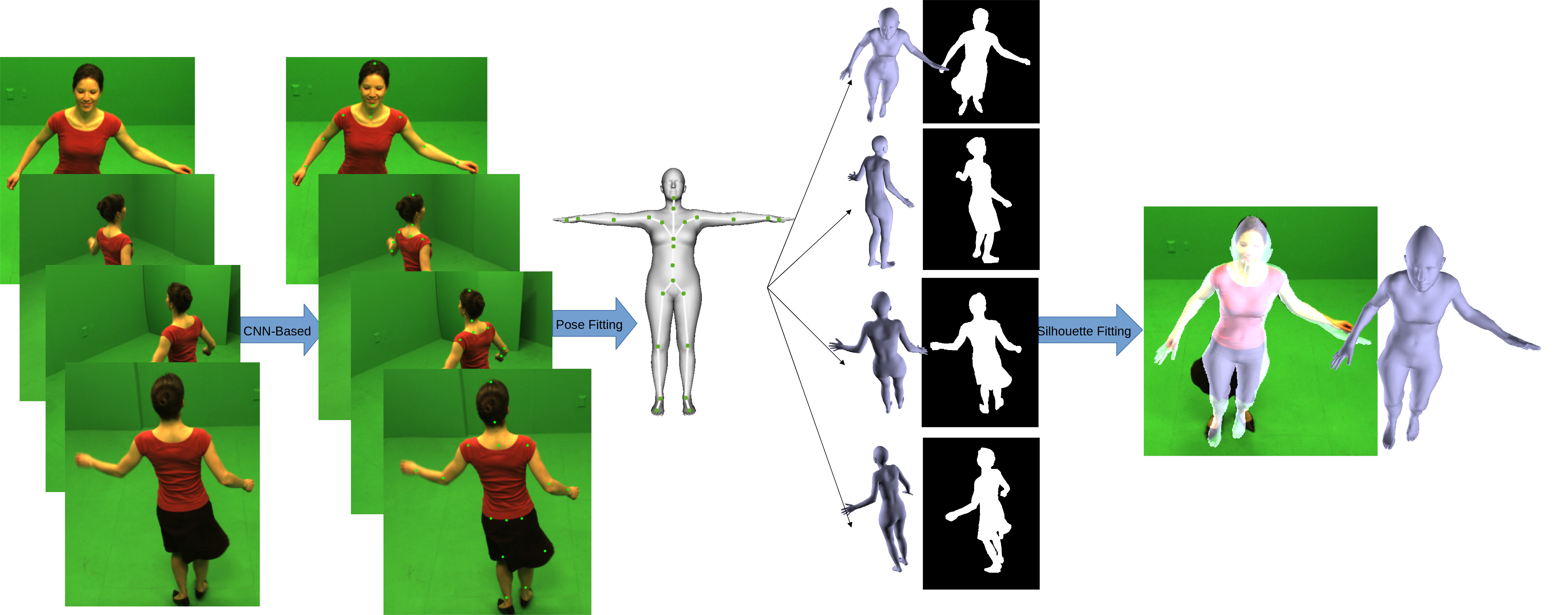}
	\caption{The overview of our method.}
	\label{fig1}
\end{figure} 
\section{Method}
In this section we present a method to obtain a 3D human body model from a small number of images taken from different view-points, using the joint points and silhouettes based on the SMPL model. The overview of the proposed method is shown in Fig.~\ref{fig1}. 
\subsection{Parametric human body model}
The parametric human body model in our method is called SMPL, and it is learned from an aligned human body dataset \cite{Loper_2015}. SMPL is defined as a function of pose $\vec{\theta}\in \mathbb{R}^{3\times 24}$ and shape $\vec{\beta} \in \mathbb{R}^{1\times 10}$ of the human body. The output of the function is a mesh with $V=6890$ vertices and $F=13776$ faces. This means that we can generate different 3D human bodies as long as we can get proper parameters of $\vec{\theta}$ and $\vec{\beta}$. There are 24 joint points in SMPL and each of them is represented as the rotation vector in terms of the root point, i.e., the $i$-$th$ joint point is represented as $\theta_i\in\mathbb{R}^{3}$. The shape parameters $\vec{\beta}$ are the first 10 coefficients of the principle components of the training dataset. 
\subsection{Pose fitting}
In the following, we explain the pose fitting in our method between SMPL and estimated joint points. For given multiple-view RGB images, the joint points are predicted by a CNN-based human pose estimation method \cite{xiao2018simple}. In order to ensure the accuracy of human pose estimation, we firstly use Cornernet \cite{Law2018} to detect the bounding box of the person, and then use the image with bounding box into \cite{xiao2018simple} to predict the joint points. Note that the order of the output of \cite{xiao2018simple} is different from the order of joints of SMPL. For given $N$ images from different views, the joint points are defined as $J_{2d}^{(i)}, i=0,...,N-1$. For the SMPL model, the joint points $J_{S}$ are in 3D space and $J_{S}$ is a function of pose $\vec{\theta}$ and shape $\vec{\beta}$. Suppose that the camera transformation matrix is $\Pi_i=(R_i,t_i)$ for the $i$-$th$ camera. The projected 2D joint points of the SMPL model on the image plane can be represented as $\Pi_i(J_{S}(\vec{\theta},\vec{\beta}))$. Therefore, the energy function to fit the SMPL model using joint points is defined as
\begin{equation}
E(\vec{\theta},\vec{\beta},\bm{R},\bm{t})=E_{jt}(\vec{\theta},\vec{\beta},R,t)+\omega_{\theta}E_{\theta}(\vec{\theta})+\omega_{\beta}E_{\beta}(\vec{\beta}) \enspace ,
\end{equation}
where $E_{jt}$ is the joint points term and $E_{\theta}(\vec{\theta}), E_{\beta}(\vec{\beta})$ are the regularization term for $\vec{\theta}, \vec{\beta}$. $\omega_{\theta}$ and $\omega_{\beta}$ are the weights of the regularization terms. $\bm{R}$ is $\{R_1, R_2, R_3\}$ and $\bm{t}$ is $\{t_1, t_2, t_3\}$. The joint points term $E_{jt}$ measures the difference between all of the joint points $J_{2d}^{(i)}$ and $\Pi_i(J_{S}(\vec{\theta},\vec{\beta}))$
\begin{equation}
E_{jt}(\vec{\theta},\vec{\beta},\bm{R},\bm{t})=\sum_{i=0}^{N-1}\rho\left(J_{2d}^{(i)}-\Pi_i(J_{S}(\vec{\theta},\vec{\beta}))\right) \enspace ,
\end{equation} 
where $\rho$ is the Geman-McClure function ~\cite{Geman_1987} and is defined as $\rho(x)=x^2/(\sigma^2+x^2)$. $\sigma$ is a constant and it is set as 100. Geman-McClure function can better deal with large noise and outliers. The regularization term for $\vec{\theta}$ is defined as
\begin{equation}
E_{\theta}(\vec{\theta})=\alpha\sum_{i=55,58,15,12}exp(\theta_i) \enspace ,
\end{equation}
where $\alpha$ is a constant which is set as 10 and the $55th$, $58th$, $15th$, and $12th$ elements in $\vec{\theta}$ are the joint points on the left and right elbows and keens. This can avoid the arms and legs to exhibit strange bending. The regularization term of $\vec{\beta}$ is defined as
\begin{equation}
E_{\beta}(\vec{\beta})=\sum_{i=0}^{9}\beta_i \enspace .
\end{equation} 
The advantage of our method is that the camera parameters are also regarded as variables. After the optimization, the rotation and translation of the cameras will also be estimated. Therefore, through the minimization of the energy function, the pose, shape parameters of the SMPL model and the camera parameters can be obtained. 
\subsection{Shape fitting}
The following section will describe the progress of shape fitting in our method. Since joint points mainly provide information about human pose in the first step, the silhouettes are used in this part to improve the estimation of shape. Here we assume that the silhouettes have been given. Now let us revisit the SMPL model about the vertex position. As shown in ~\cite{Loper_2015}, the vertex of SMPL is transformed as
\begin{equation}
\bm{t}_i=\sum_{k=1}^{K}\omega_{k,i}G'_k(\vec{\theta},J(\vec{\beta}))\left(\bar{\bm{t}}+B_S(\vec{\beta})+B_P(\vec{\theta})\right) \enspace .
\end{equation}
\begin{wrapfigure}{r}{0.3\textwidth}
\centering
	\includegraphics[width=0.8in]{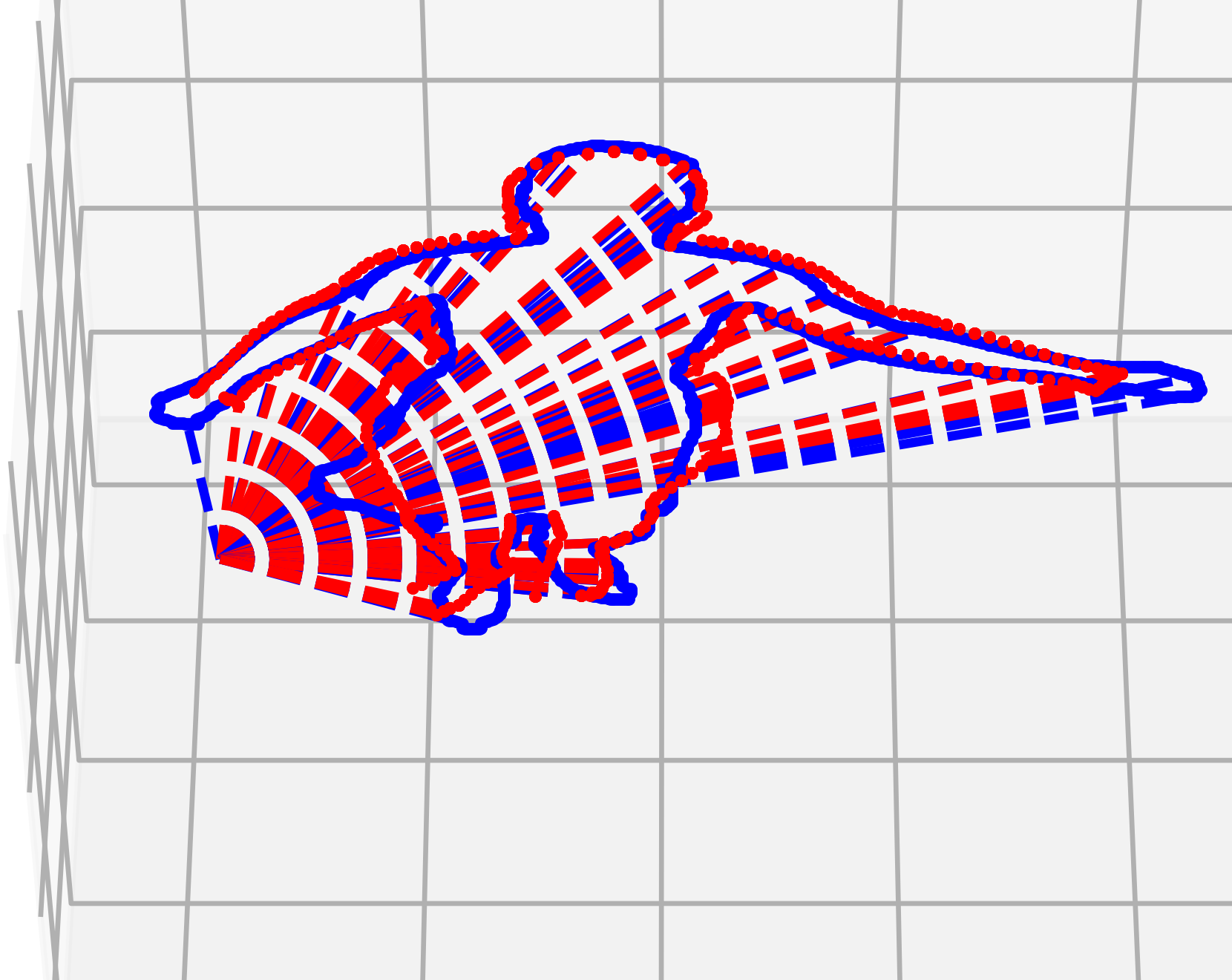}%
	\includegraphics[width=0.6in]{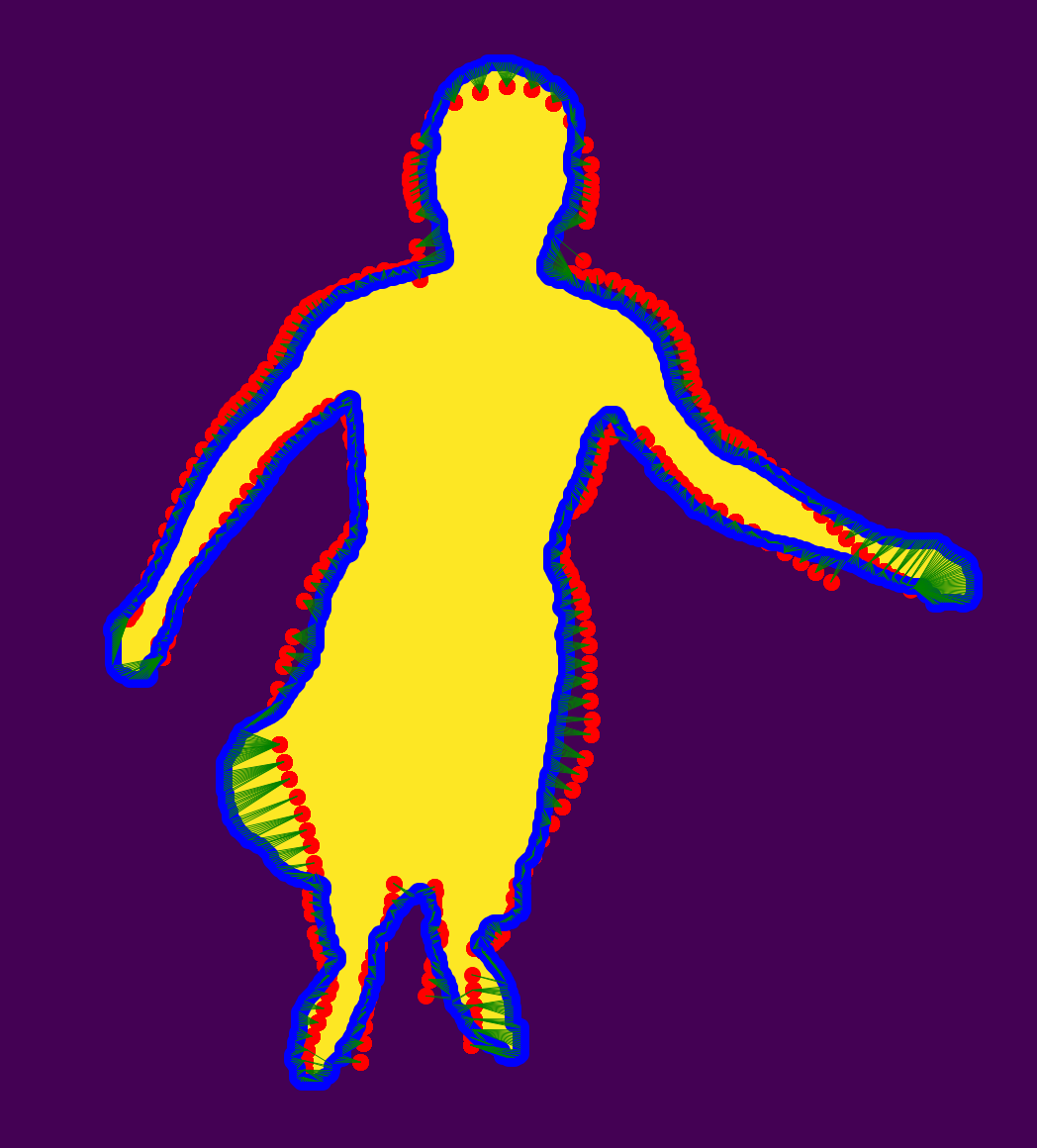}
	\caption{An example of correspondence between silhouette and SMPL model in 2D and 3D space. The left is the 3D correspondence and the right is the 2D correspondance between SMPL and silhouettes. The red points are SMPL vertices and the blue points are the corresponding points on silhouettes.}
	\label{fig2}
\end{wrapfigure}
In addition, since the rotation $R$ and translation $t$ of the camera are estimated after pose fitting, the positions of the cameras can be computed as $c=-R^Tt$. Thus, we can define a ray from the camera $c$ to the vertex $\bm{t}_i$ of the transformed SMPL model as $\vec{r}$, as shown in Fig.~\ref{fig2}. Then, for the untransformed SMPL model, the corresponding ray is 
\begin{equation}
\vec{r}'=\left[\sum_{k=1}^{K}\omega_{k,i}G'_k(\vec{\theta},J(\vec{\beta}))\right]^{-1}\vec{r}-B_P(\vec{\theta}) \enspace .
\end{equation}
We would like to find the correspondence between $\vec{r}'$ and the boundary points of the observed silhouette. This ray can be decomposed using Plucker coordinates $(\vec{r}'_m,\vec{r}'_n)$. Given the silhouette of the image, we can find the boundary points $\bm{v}$ of the silhouette and then backproject $\bm{v}$ to $\bm{V}$ in the camera coordinates since we have estimated the camera parameters. Then, the distance from the points to the ray can be computed as $d=\bm{V}\times \vec{r}'_n -\vec{r}'_m$. Those points and rays whose distance is smaller then a threshold are regarded as corresponding pairs. These pairs are defined as a set $P$ which is the correspondence in 3D space. On the other hand, the vertices of SMPL model intersected by ray $\vec{r}'$ can be projected to the image plan as $\bm{v}'$ using camera parameters. The point set $\bm{v}$ and $\bm{v}'$ are defined as $Q$, which is the correspondence in 2D space. Fig.~\ref{fig2} shows one example of the correspondence on the SMPL vertices and the silhouettes in 2D and 3D space. We can see that the correspondence in this case seems to be correct and can provide additional information for the shape fitting. Overall, the energy function using silhouettes is defined as
\begin{equation}
E(\vec{\beta})=E_{silh}(\vec{\beta})+E_{reg}(\vec{\beta}) \enspace .
\end{equation}
The silhouette term $E_{silh}(\vec{\beta})$ is constructed by using the set $P$ and $Q$ and is defined as
\begin{equation}
E_{silh}(\vec{\beta})=\sum_{(\bm{V},\vec{r})\in P}\rho(\bm{V}\times \vec{r}'_n -\vec{r}'_m)+\sum_{(\bm{v},\bm{v}')\in Q}\rho(\bm{v}-\bm{v}') \enspace ,
\end{equation}
where $\bm{V}\times \vec{r}'_n$ is the cross product of $\bm{V}$ and $\vec{r}'_n$, $\rho$ is the Geman-McClure function as Eq.(2). The first part of $E_{silh}$ measures the difference of 3D points of backprojected silhouette boundary and rays, while the second part shows the difference of 2D silhouette points and projected SMPL vertices. Therefore, the silhouette term considers the silhouette information from both 3D and 2D perspective in contrast to the paper \cite{alldieck_2018}.

The regularization term is defined based on the SMPL model with zero pose, i.e., $\vec{\theta}=\vec{0}$. This is because this part only focuses on the shape estimation. Then, the SMPL model is computed as $\bm{t}(\vec{\beta},D)=\bar{t}+B_S{\vec{\beta}}+D$, where $D$ is the offset given by the SMPL model. The regularization term contains the Laplacian term $E_L$ as well as the body model term $E_B$ and it is represented as in \cite{alldieck_2018}
\begin{equation}
E_{reg}(\vec{\beta})=\omega_{L}E_L+\omega_{B}E_B \enspace ,
\end{equation} 
where $\omega_{L}$ and $\omega_{B}$ are the weights. The Laplacian term $E_L$ is defined as
\begin{equation}
E_L=\sum_{i=1}^{N}||L(\bm{t}_i)-\delta_i||^2 \enspace ,
\end{equation}
where $L$ is the Laplace operator and $\delta_i=L(\bm{t}_i(\vec{\beta},0))$. This term enforces smooth deformation.
The body model term $E_B$ is represented as
\begin{equation}
E_B=\sum_{i=1}^N||\bm{t}_i(\vec{\beta},D)-\bm{t}_i(\vec{\beta},0)||^2 \enspace .
\end{equation}
Through minimizing (7), the shape parameters can be estimated and the final results are obtained.
\subsection{Optimization}
After building the energy functions based on joint points and silhouettes, we need to optimize the energy functions. We have used Python to implement our optimization method. The energy functions in (1) and (7) are minimized by Powell’s dogleg method which is provided in the Python modules called OpenDR \cite{Loper_2014} and Chumpy. For four images with different views, it takes about 2 minutes to obtain the final estimation of the 3D human body. 
\begin{table}[htbp]
\caption{The values of parameters for the optimization.}\label{tab1}
\centering
\begin{tabular}{|c|c|c|c|c|c|c|c|c|c|c|c|c|}
\hline 
 &\multicolumn{6}{|c|}{Synthtic dataset} & \multicolumn{6}{|c|}{Real dataset}\\
\hline 
 &\multicolumn{3}{|c|}{Pose fitting} & \multicolumn{3}{|c|}{Shape fitting} & \multicolumn{3}{|c|}{Pose fitting} & \multicolumn{3}{|c|}{Shape fitting} \\
\hline 
k & $\omega_{\theta}$ & $\omega_{\beta}$ & $\sigma$ & $\omega_L$ & $\omega_B$ & $\sigma$ & $\omega_{\theta}$ & $\omega_{\beta}$ & $\sigma$ & $\omega_L$ & $\omega_B$ & $\sigma$ \\
\hline
1 & 91.0 & 100 & 100 & 6.5 & 0.9 & 0.05 & 91.0 & 100 & 100 & 6.5 & 0.9 & 0.08 \\
2 & 91.0 & 50  & 100  & 5.25& 0.75& 0.03 & 91.0 & 50  & 100 & 5.25  & 0.75& 0.04 \\
3 & 47.4 & 10  & 100 & 4  & 0.6 & 0.01 & 47.4 & 10 & 100 & 4 & 0.6 & 0.03 \\
4 & 4.78 & 5   & 100 &    &     &      & 4.78 & 5  & 100 &  &     &  \\
\hline
\end{tabular}
\end{table}
The parameters used during the optimization are shown in Table 1. In the following experiments, we mainly used a synthetic dataset and a real dataset to evaluate our approach. Table 1 gives the parameters that we used in the experiments for the two datasets. For pose fitting, we assume that the focal length of the camera is known, but the translation and rotation of the camera are unknown. We initialize the rotation matrix as the identity matrix. The translation vector is initialized according to the torso length of SMPL model and the torso length of human body in the images. The weights $\omega_{\theta}$ and $\omega_{\beta}$ in the energy function are decreased gradually after some iterations or when the value of the energy function is smaller than a threshold. For the silhouettes based energy function, we assume that the ground truth of the silhouettes are given. The weights $\omega_L$, $\omega_B$ and the parameter $\sigma$ in Geman-McClure function are decreased gradually after some iterations or when the value of the function is smaller than a threshold. 
\section{Experiments}
In this section, the experiments to evaluate our proposed method are presented. We firstly introduced the datasets which were used in the experiments. Then, we discussed the effect of joint points on pose fitting and the influence of silhouettes on shape fitting, respectively. Besides, we also evaluated the pose fitting and shape fitting on the final estimation. Finally, we compared our method to several previous approaches on the datasets to validate the advantage of our approach.  
\subsection{Datasets}
To evaluate our approach for a variety of poses and shapes, we generated a synthetic dataset and also used a public real dataset. The synthetic dataset consisted of 50 male and 50 female 3D human bodies which were created by the SMPL model. We set all the human bodies as "A" pose through giving the same pose parameters of the SMPL model, while the shape of each human body was different by varying the shape parameters of the SMPL model. For each 3D human body, we used four cameras from different views to project the 3D model into four 2D images. Since the 3D joint points of the SMPL model relied on the pose and shape parameters, the ground truth of 2D joint points and silhouettes can also be obtained when we projected the SMPL model. 

The real dataset is from \cite{Vlasic2008} and consists of ten image sequences. Each sequence was captured by eight cameras in an indoor scene. Four images which are taken by the 2nd, 4th, 6th and 8th cameras are adopted in our experiments. Note that there are two marches and squats in the dataset, so we evaluate the results of march1 and squat2, i.e, the experiments are implemented based on eight image sequences. We predicted the bounding box the person through Cornernet \cite{Law2018}, and then estimated the joint points of the dataset through the method in \cite{xiao2018simple}. In terms of the silhouettes, the ground truth was given in this dataset. However, the silhouettes can be extracted through threshold and filter since the background can be easily removed. In practice, semantic segmentation can be used for silhouettes extraction. Silhouette segmentation is not the key problem in our method, 
So we directly use the ground truth of silhouettes like \cite{alldieck_2018}. 

The metric for quantitatively comparison is the intersection over union (IoU) between the ground truth (GT) of silhouettes and the silhouettes of the estimated human body model from multiple views. Note that the GT of silhouettes here is used for computing the IoU, which has no relationship with the silhouettes in the optimization. This is because the silhouettes for optimzation can use the GT of silhouettes or be obtained by some other segmentation algorithms.
\subsection{Evaluation of pose fitting and shape fitting}
\begin{figure}[t]
	\centering
	\begin{minipage}{0.49\linewidth}
    \centering
    \includegraphics[width=0.55in, height=0.6in]{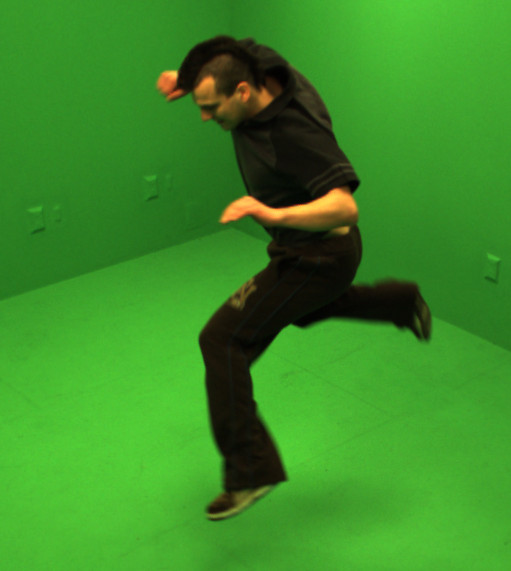}%
	\includegraphics[width=0.55in, height=0.6in]{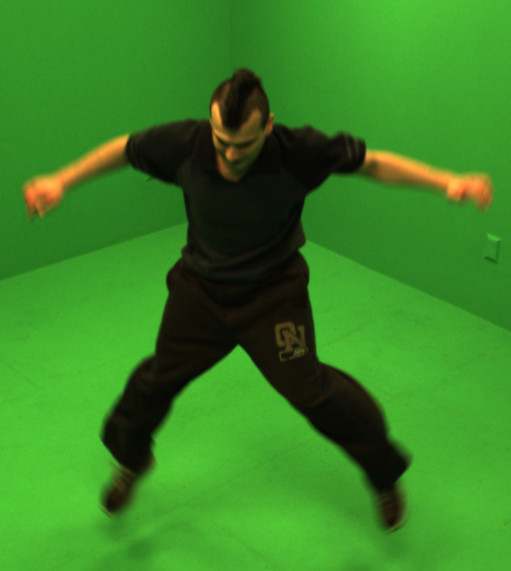}%
	\includegraphics[width=0.55in, height=0.6in]{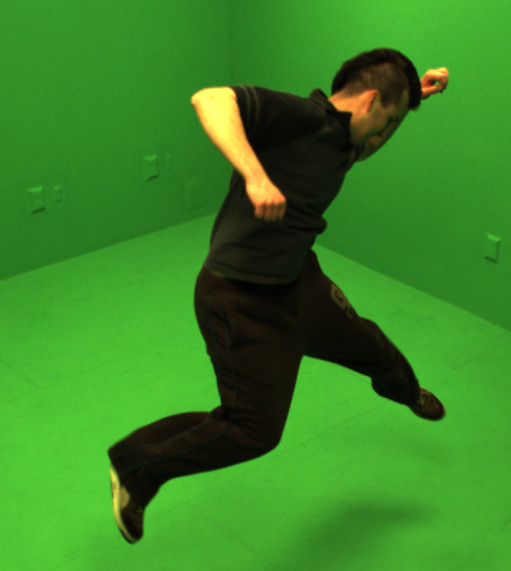}%
	\includegraphics[width=0.55in, height=0.6in]{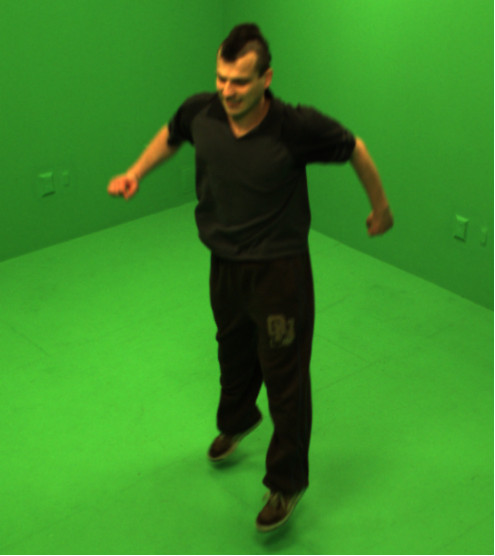}
		
	\includegraphics[width=0.55in, height=0.6in]{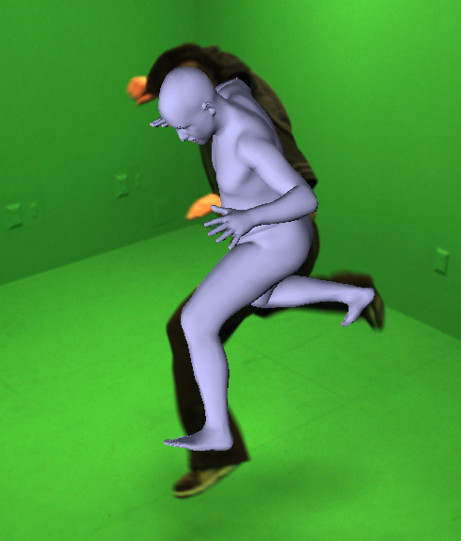}%
	\includegraphics[width=0.55in, height=0.6in]{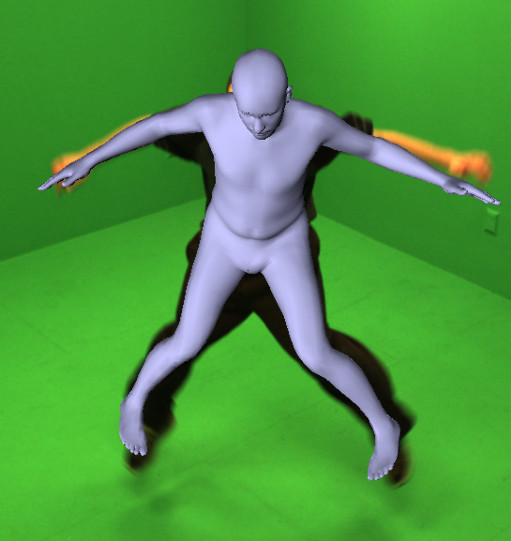}%
	\includegraphics[width=0.55in, height=0.6in]{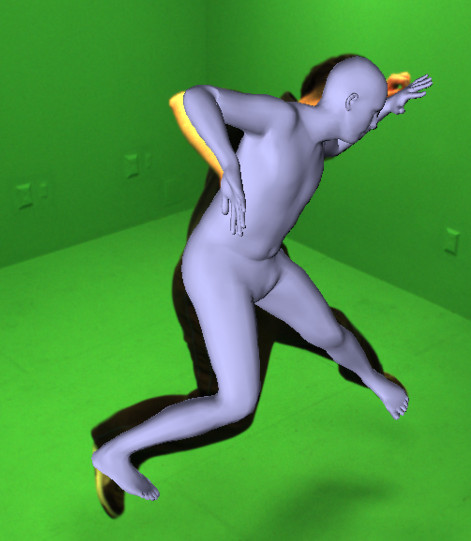}%
	\includegraphics[width=0.55in, height=0.6in]{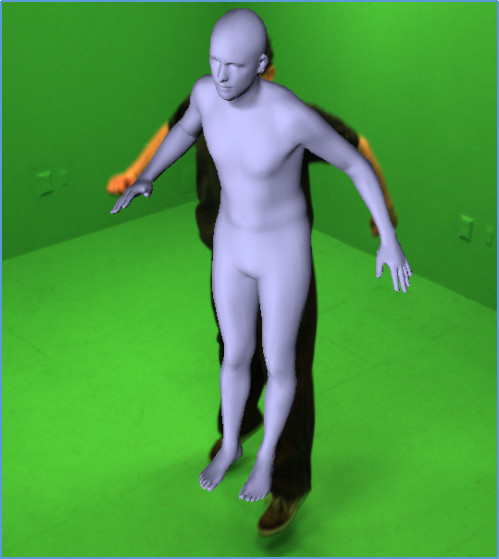}
	\includegraphics[width=0.55in, height=0.6in]{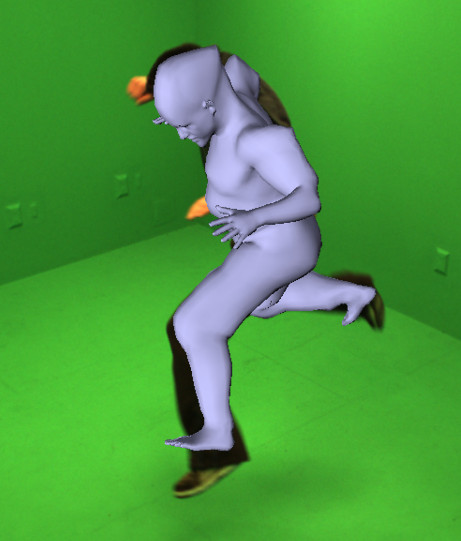}%
	\includegraphics[width=0.55in, height=0.6in]{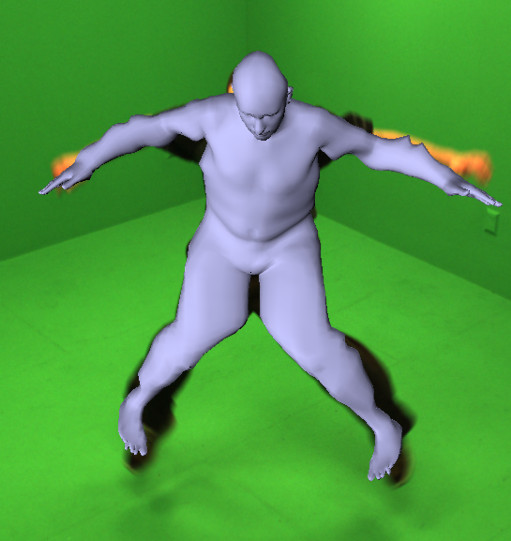}%
	\includegraphics[width=0.55in, height=0.6in]{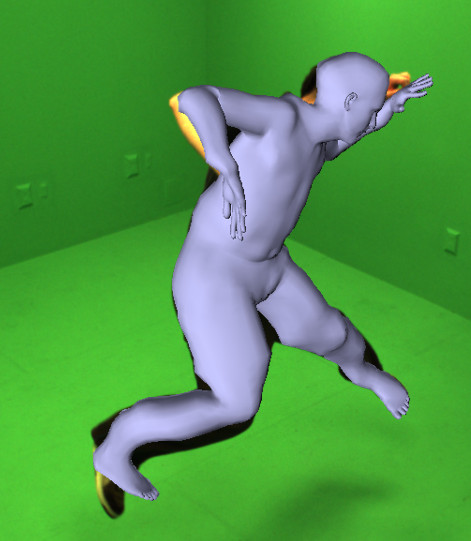}%
	\includegraphics[width=0.55in, height=0.6in]{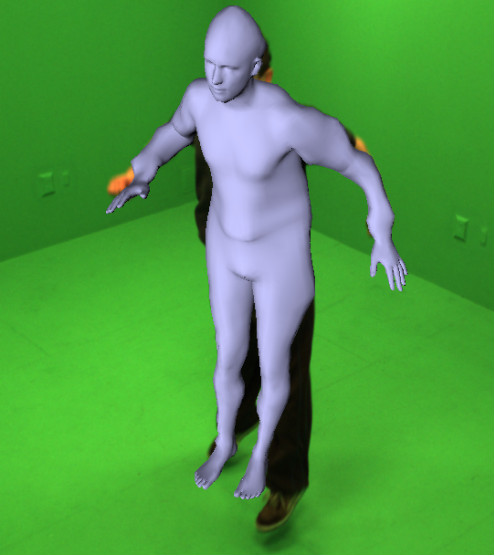}%
	\caption{Comparison of pose fitting and shape fitting. From top to down: Original images, results after pose fitting and results after shape fitting. From left to right: the 64th, 85th, 115th and 145th frames.}
	\label{fig_r1}
    \end{minipage}
    \begin{minipage}{0.49\linewidth}
    \centering
    \includegraphics[width=\textwidth]{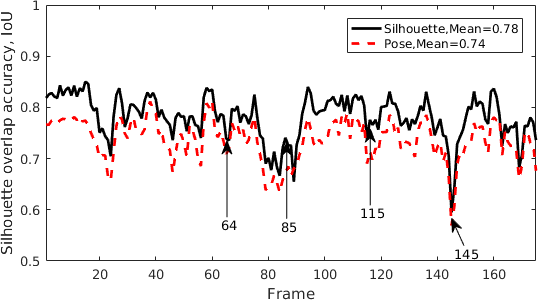}
		\caption{Quantitative comparison of IoU of the pose fitting and shape  fitting on the \textit{Bouncing}.}
	\label{fig_r2}
    \end{minipage}
\end{figure}%

We use the real data to evaluate the performance of pose fitting and shape fitting on the final results. The qualitative and quantitative results of pose fitting and shape fitting for \textit{Bouncing} from the real dataset are shown in Fig.~\ref{fig_r1} and Fig.~\ref{fig_r2}, respectively. The results in Fig.~\ref{fig_r1} are the frames of 64, 85,115 and 145, which is pointed out in Fig.~\ref{fig_r2}. We can see that the human bodies after shape fitting are better fitted to the original images. Even for the 145th frame which has the lowest IoU due to the effect of pose fitting, the final human body is better than the result only relying pose fitting. The IoU of silhouettes from four views of the image sequence \textit{Bouncing} are shown in Fig.~\ref{fig_r2}. It is shown from the figure that the IoU after silhouette fitting is higher than the IoU only using pose fitting for most frames in the sequence. The mean IoU after shape fitting is 0.78, while the mean IoU only using pose fitting is 0.74, which shows that shape fitting is a step to improve the accuracy of human body reconstruction. 

\subsection{Comparison to previous approaches}
\begin{figure}
	\centering
	\includegraphics[width=0.96in]{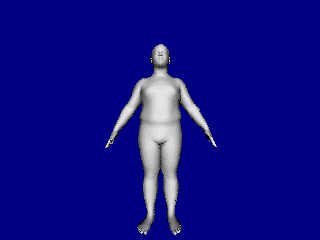}%
	\includegraphics[height=0.72in]{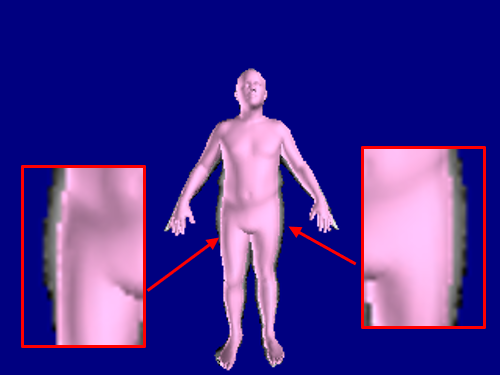}%
	\includegraphics[height=0.72in]{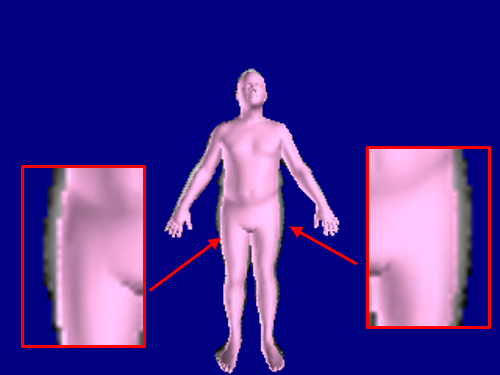}%
	\includegraphics[height=0.72in]{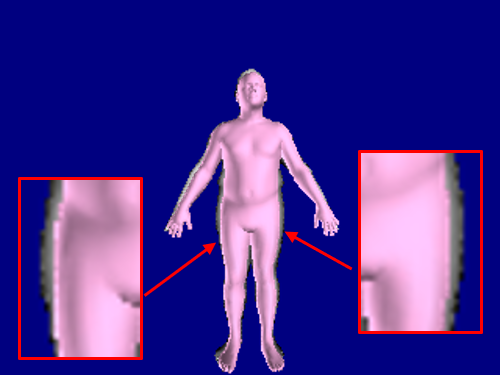}%
	\includegraphics[height=0.72in]{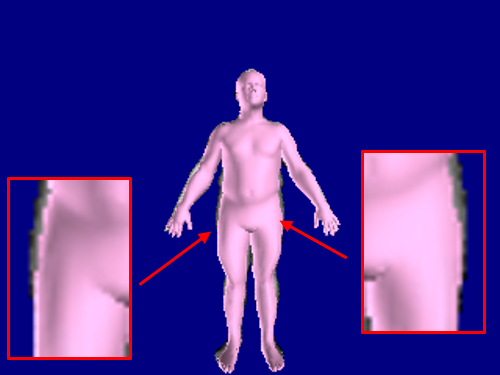}%
	\hspace{0.1in}
	\includegraphics[width=0.96in]{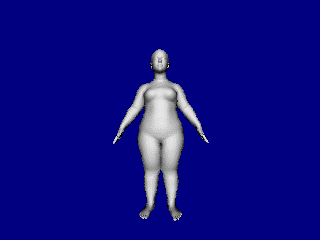}%
	\includegraphics[height=0.72in]{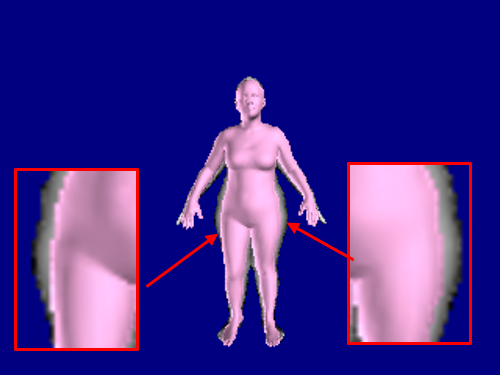}%
	\includegraphics[height=0.72in]{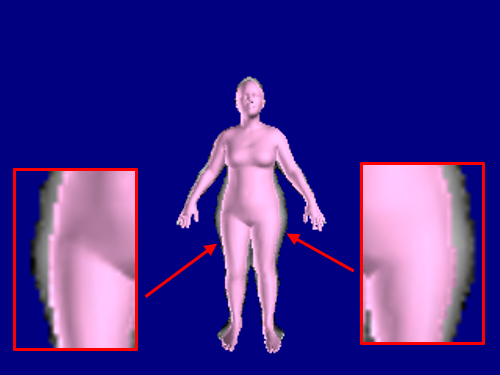}%
	\includegraphics[height=0.72in]{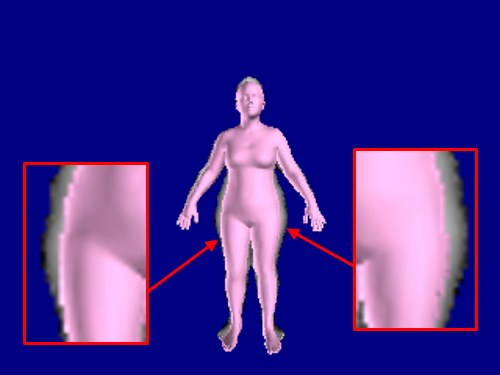}%
	\includegraphics[height=0.72in]{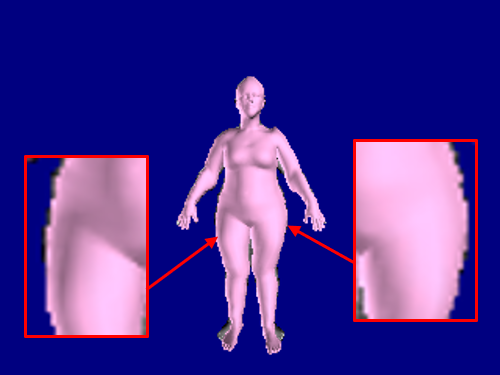}
	\caption{Two results on the synthetic data of one view. From left to right: the original images, SMPLify \cite{Bogo_2016}, SMPLify4 \cite{Li_2019}, VideoAvatar \cite{alldieck_2018} and our method. }
	\label{fig8}
\end{figure}

\begin{figure}
\centering
\subfigure[Male]{
\begin{minipage}[b]{0.48\linewidth}
\includegraphics[width=0.33\textwidth]{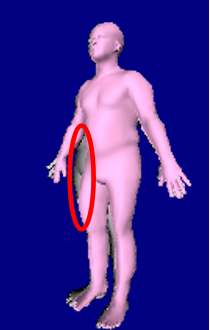}%
\includegraphics[width=0.33\textwidth]{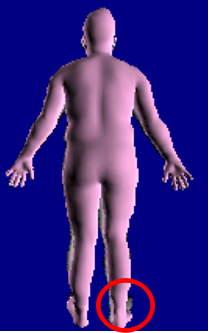}%
\includegraphics[width=0.33\textwidth]{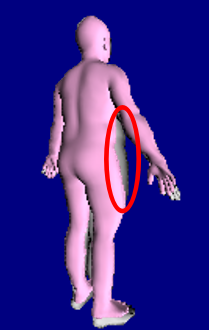}
\end{minipage}}
\subfigure[Female]{
\begin{minipage}[b]{0.48\linewidth}
	\includegraphics[width=0.33\textwidth]{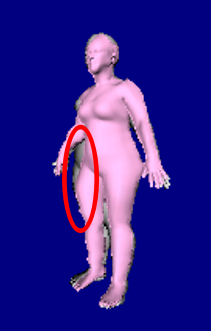}%
	\includegraphics[width=0.33\textwidth]{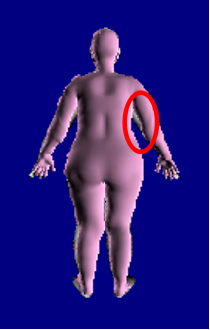}%
	\includegraphics[width=0.33\textwidth]{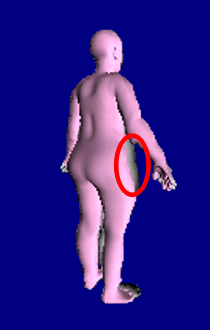}
\end{minipage}}
\caption{The results of the other three views obtained by our method.}
\label{fig9}
\end{figure}

In this section we evaluate our method on both the synthetic and real dataset. We compared to three previous approaches: SMPLify \cite{Bogo_2016}, SMPLify4 \cite{Li_2019} and VideoAvatar \cite{alldieck_2018}. Fig.~\ref{fig8} qualitatively shows the comparison of two examples from the synthetic data. The areas indicated by red rectangles shows that our method can better recover the shape of human body model than the other three methods. The estimated human body models (pink models) by our method better fitted to the original human body models (white models). Especially for VideoAvatar, since our method established the energy function of silhouettes in 2D and 3D space, the results using four images are better than VideoAvatar which used 120 images from different view-points to obtain the 3D human body model. Fig.~\ref{fig9} shows the results of our method from the other three view-points of the male and female model in Fig.~\ref{fig8}. Although the areas indicated by read circles are still not accurate enough, the results are satisfying.

\begin{figure}
\begin{minipage}{0.5\linewidth}
\centering
\includegraphics[width=\textwidth]{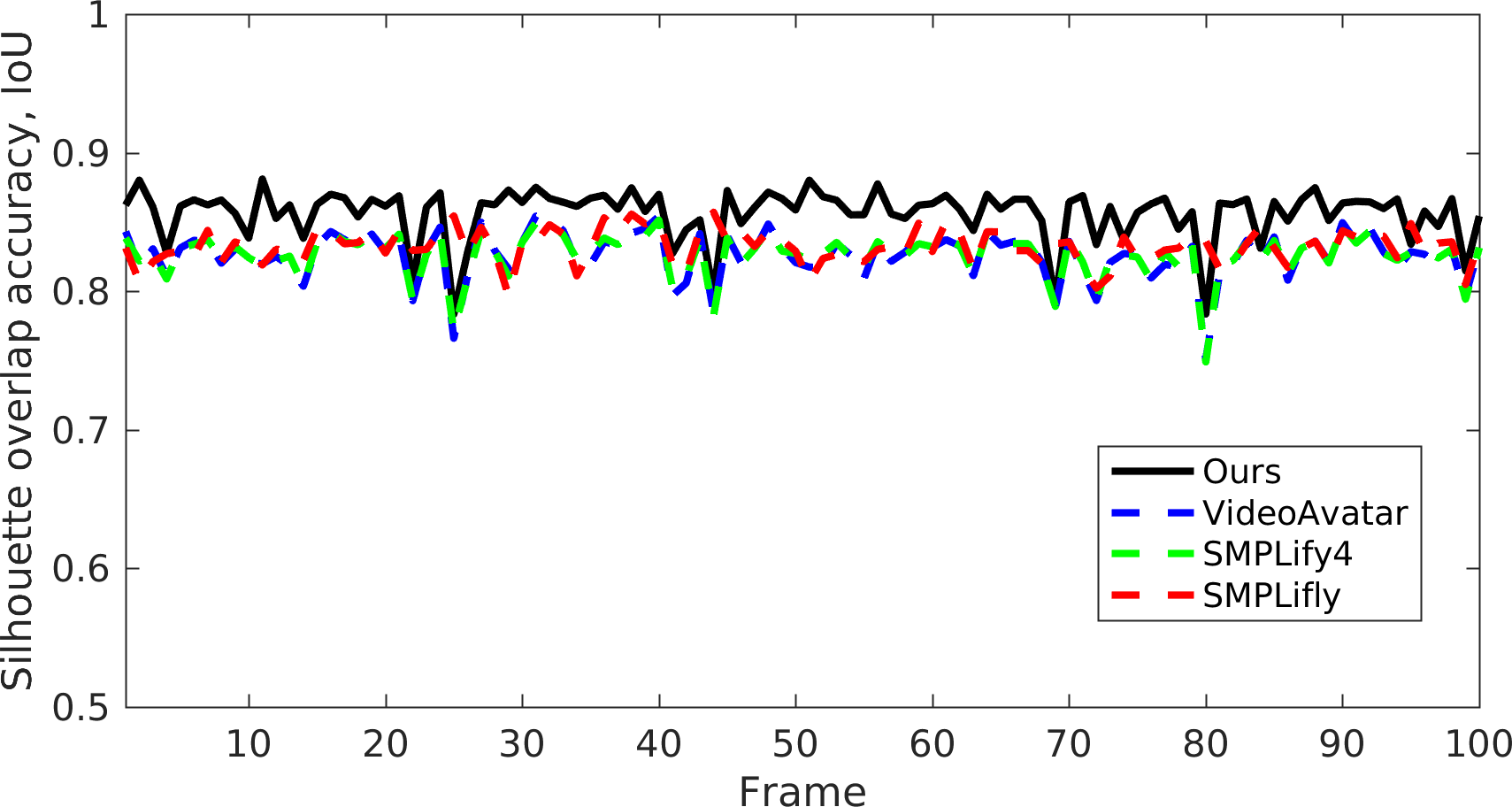}
\caption{The comparison of IoU of silhouette overlap between our method and other methods on the synthetic dataset.}
\label{fig10}
\end{minipage}
\begin{minipage}{0.5\linewidth}
\centering
\includegraphics[width=\textwidth]{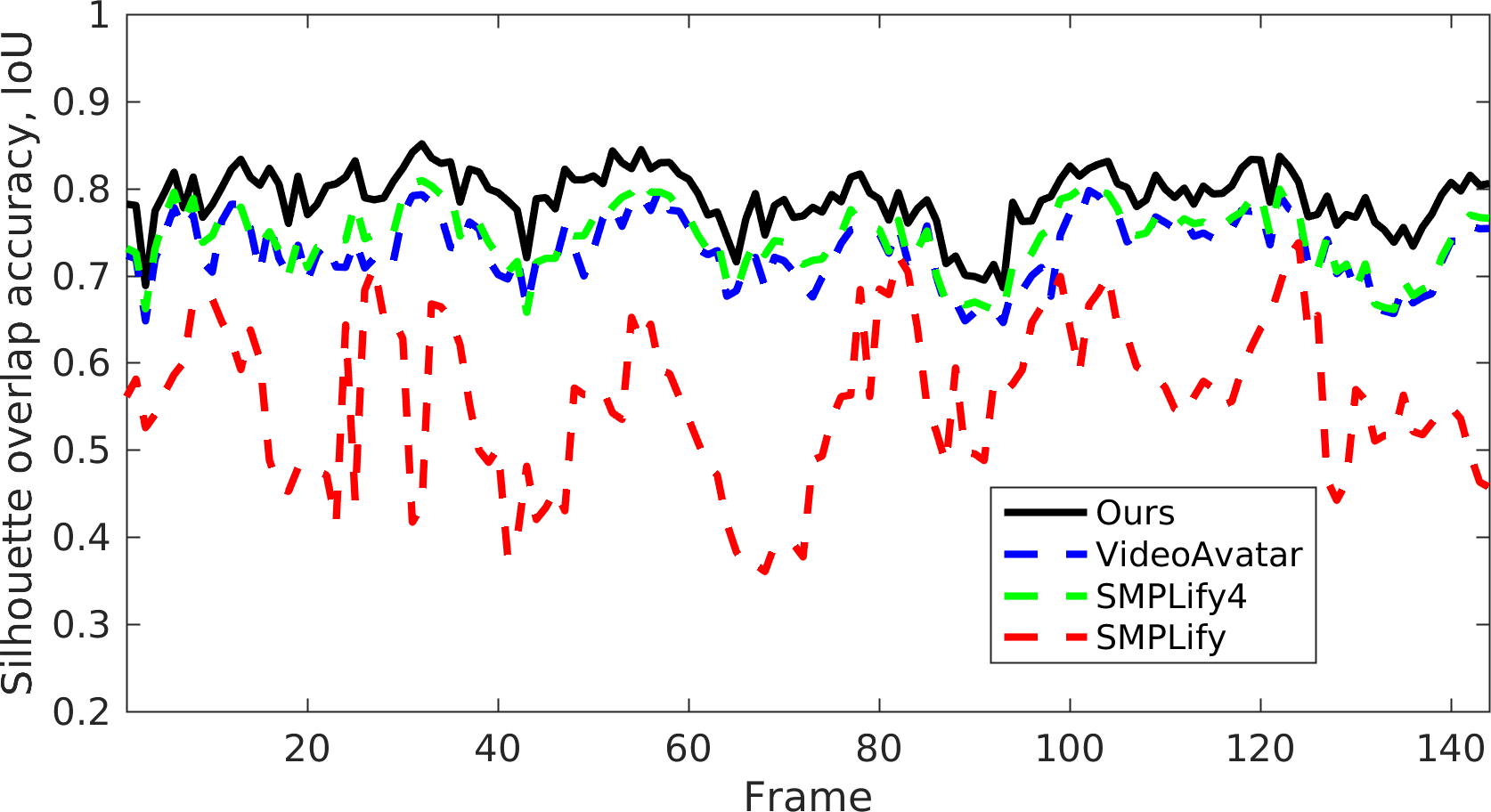}
\caption{The comparison of IoU of silhouette overlap between our method and other methods on the the \textit{Crane} image sequence in read dataset.}
\label{fig11}
\end{minipage}
\end{figure}

Fig.~\ref{fig10} shows the IoU of silhouette overlap of our methods compared to the other three methods on the synthetic dataset. The IoU of silhouette overlap is computed based on the silhouette of the projected 3D human body model and the corresponding ground truth of silhouette. The IoU for SMPLify is calculated based on one view-point, while the IoU of the other methods is based on four view-points. Although the results of some samples for SMPLify are better, the accuracy of our method is still higher than the results of the other three methods for the most samples in the dataset. Since SMPLify only adopts one image, the optimization is not sensitive to the initialization, which is the reason that the results of SMPLify on some samples are better. Compared to the SMPLify4 \cite{Li_2019}, our method introduced silhouette after the joint points optimization, and thus, the results of our method are better on the synthetic dataset. In addtion, for VideoAvatar \cite{alldieck_2018} which also uses silhouettes, our results are still better. The reason could be that the energy function of silhouette in VideoAvatar was only built from 3D space. In VideoAvatar, the 3D model was acquired from a video stream containing 120 frames from multiple viewpoints. By contrast, the improved energy function of silhouettes in our method considers both 2D and 3D, which ensures that our method have good performance only using four images.

In the following, we evaluate our method on the data in \cite{Vlasic2008}. Firstly, we show the IoU of silhouette overlap for image sequence \textit{Crane} in Fig.~\ref{fig11} by comparing to SMPLify \cite{Bogo_2016}, SMPLify4 \cite{Li_2019}, VideoAvatar \cite{alldieck_2018}. Note that the IoU of SMPLify is also computed based on four images to reflect the accuracy of the 3D models. 
We can see that our method obtains higher accuracy than the other three methods for the most samples in this image sequence. The results of SMPLify are the worst because only joint points from one single image are used as prior information. The results of SMPLify4 and VideoAvarta are almost the same because VideoAvatar requires enough number of images from different views. Furthermore, we also give the average of IoU of silhouettes overlap for the eight different actions in the real dataset in Table 2. It is shown from this table that our method achieves the best performance comparing to other three previous methods because the IoU of our method is the highest. The results of SMPLify are worst, while the SMPLify4 and VideoAvatar have almost the same performance and they are better than SMPLify. The results of \textit{Handstand} are not good because the pose estimation for the images in the sequence is not good. The pretrained model in human pose estimation of \cite{xiao2018simple} cannot achieve good estimation for human body with handstand. Even in this case our results are sill the best comparing to other methods. Overall, our method has competitive performance among these approaches according to Table 2.

\begin{figure}[H]
	\centering
	\subfigure[Original]{
		\begin{minipage}[b]{0.19\linewidth}
		\includegraphics[width=1\linewidth ,height=1.2\linewidth]{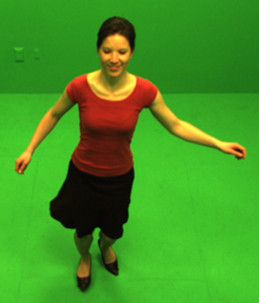}
		\includegraphics[width=1\linewidth, height=1\linewidth]{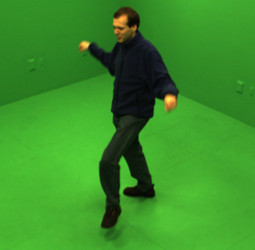}
		\includegraphics[width=1\linewidth ,height=1\linewidth]{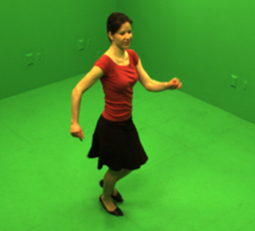}
		\includegraphics[width=1\linewidth ,height=1.2\linewidth]{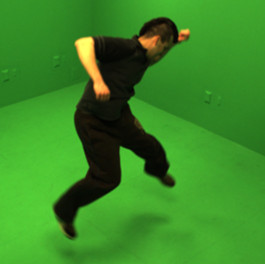}
		\end{minipage}}%
	\subfigure[SMPLify ]{
		\begin{minipage}[b]{0.19\linewidth}
		\includegraphics[width=1\linewidth ,height=1.2\linewidth]{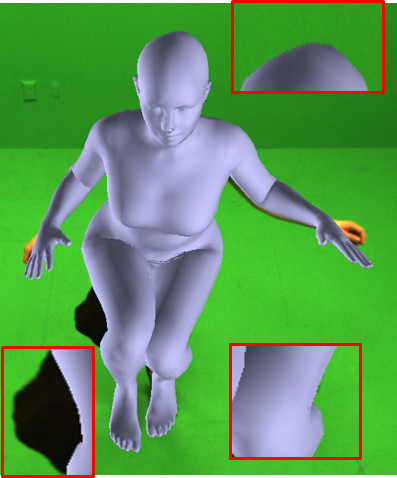}
		\includegraphics[width=1\linewidth ,height=1\linewidth]{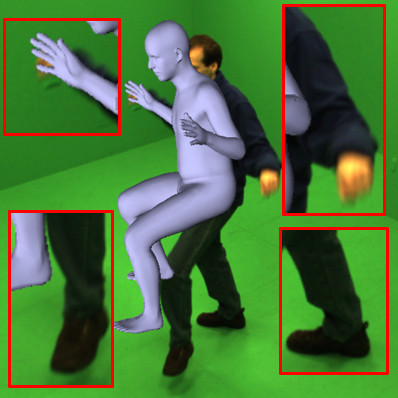}
		\includegraphics[width=1\linewidth ,height=1\linewidth]{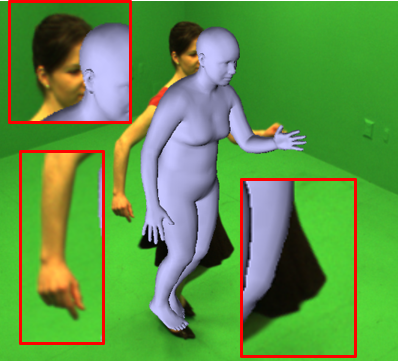}
		\includegraphics[width=1\linewidth ,height=1.2\linewidth]{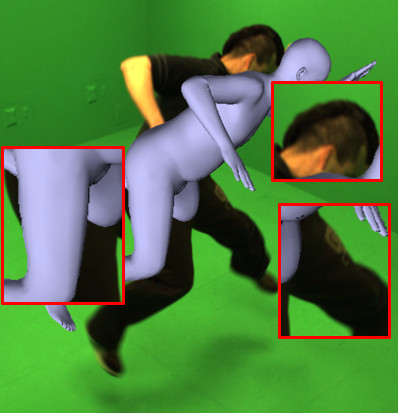}
		\end{minipage}}%
	\subfigure[ SMPLify4]{
		\begin{minipage}[b]{0.19\linewidth}
		\includegraphics[width=1\linewidth ,height=1.2\linewidth]{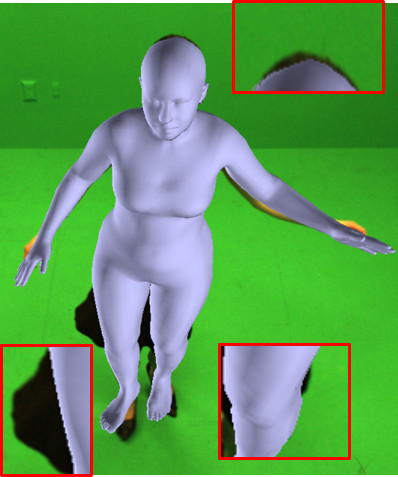}
		\includegraphics[width=1\linewidth ,height=1\linewidth]{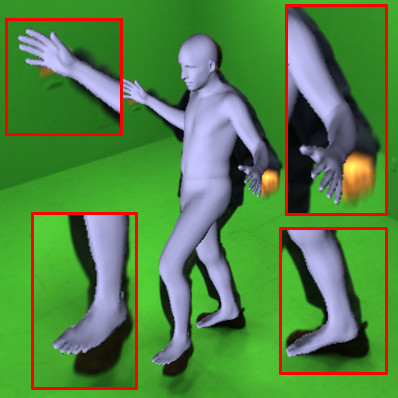}
		\includegraphics[width=1\linewidth ,height=1\linewidth]{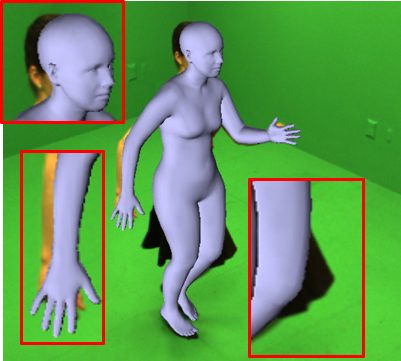}
		\includegraphics[width=1\linewidth ,height=1.2\linewidth]{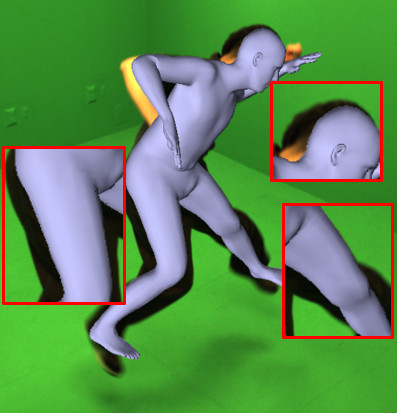}
		\end{minipage}}%
	\subfigure[VideoAvatar]{
		\begin{minipage}[b]{0.19\linewidth}
		\includegraphics[width=1\linewidth ,height=1.2\linewidth]{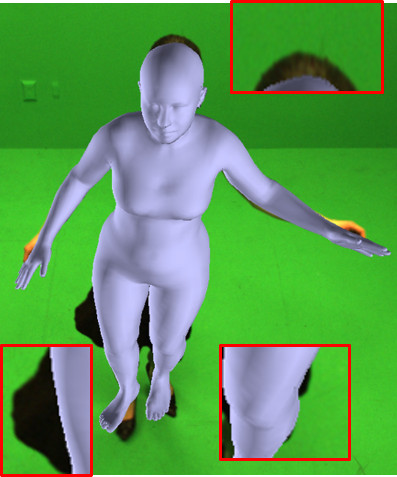}
		\includegraphics[width=1\linewidth ,height=1\linewidth]{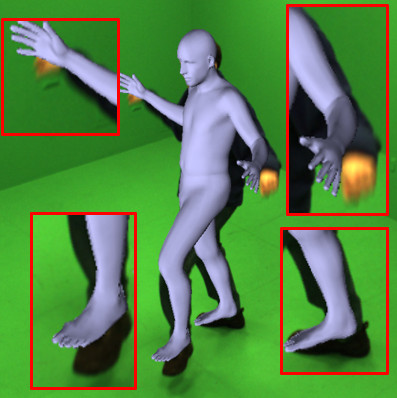}
		\includegraphics[width=1\linewidth ,height=1\linewidth]{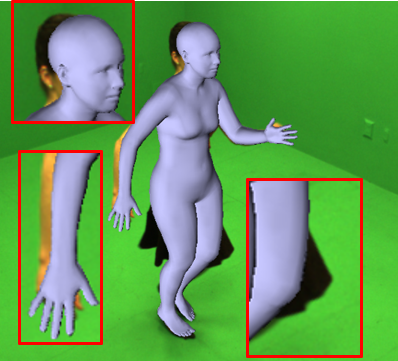}
		\includegraphics[width=1\linewidth ,height=1.2\linewidth]{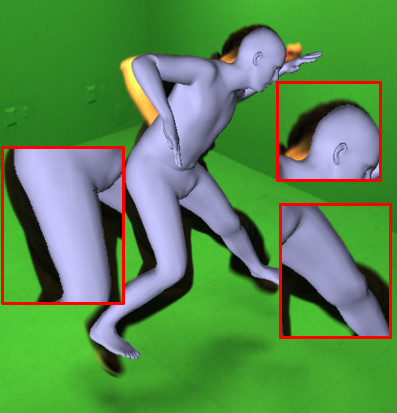}
		\end{minipage}}%
	\subfigure[Proposed]{
		\begin{minipage}[b]{0.19\linewidth}
		\includegraphics[width=1\linewidth ,height=1.2\linewidth]{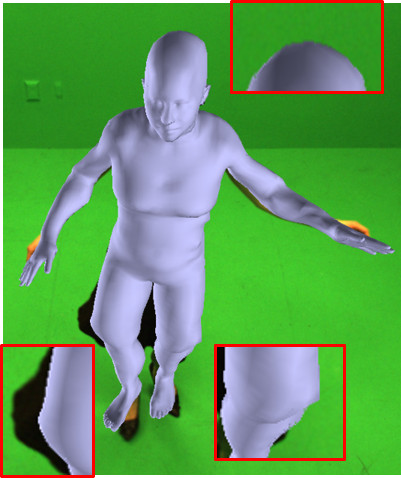}
		\includegraphics[width=1\linewidth ,height=1\linewidth]{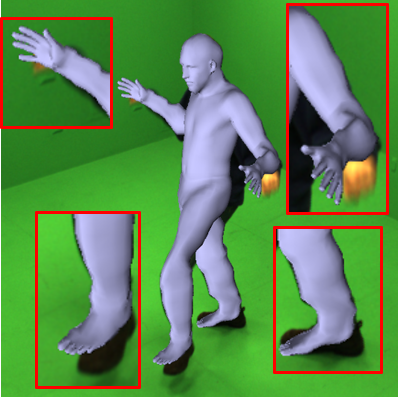}
		\includegraphics[width=1\linewidth ,height=1\linewidth]{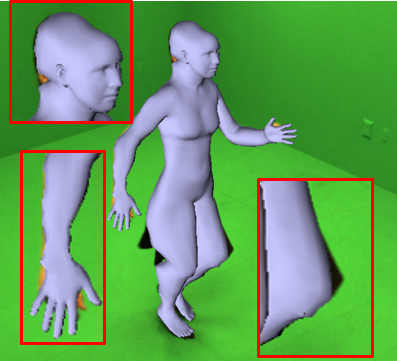}
		\includegraphics[width=1\linewidth ,height=1.2\linewidth]{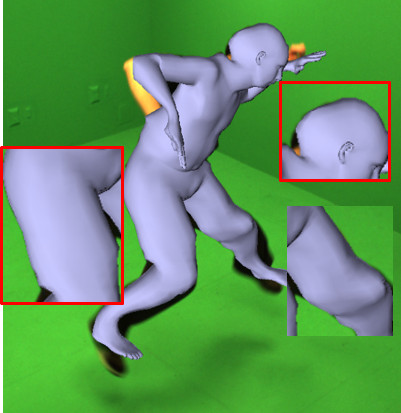}
		\end{minipage}}
	\caption{The results of \textit{Swing, Crane, Samba and Bouncing} from top to down. The original images and the results of SMPLify \cite{Bogo_2016}, SMPLify4 \cite{Li_2019}, VideoAvatar \cite{alldieck_2018} and proposed method are shown in (a), (b), (c), (d) and (e). }
	\label{fig12}
\end{figure}

Several images from \textit{Swing}, \textit{Crane}, \textit{Samba} and \textit{Bouncing} are shown in Fig.~\ref{fig12}. In the figure, we illustrate one frame from each above sequence and give the qualitative results of the three previous methods and our method from one view. We can see from the figure that the shape of our method gives a better fit to the original images, which can be seen from the parts that are zoomed in. More specifically, the \textit{Bouncing} results of SMPLify are not correct. This is because using only a single RGB image gives a too high uncertainty concerning the spatial information. Compared to SMPLigy4 and VideoAvatar, which are shown in the second and third columns in Fig.~\ref{fig12}, the shapes of our method provides a better fit because to the original images. This demonstrates the effectiveness to use the energy function based on silhouettes from 2D and 3D space. Therefore, our method achieves a good estimation not only for the pose but also for the shape of the human body. We also provide the results obtained by our method from the other three views in Fig.~\ref{fig12}. It demonstrates that the results from other views are correct, which means that the 3D model estimated by our method has better accuracy.

\begin{table}[htbp]
\caption{The mean IoU of silhouette overlap of the 8 image sequence in the real dataset.}\label{tab2}
\centering
\begin{tabular}{|l|l|l|l|l|l|}
\hline
 &  Frames & SMPLify \cite{Bogo_2016} & SMPLify4 \cite{Li_2019}  & VideoAvatar \cite{alldieck_2018} & Ours \\
\hline
Swing   & 150 & 0.5649 & 0.7570 & 0.7573 & \textbf{0.7748} \\
Crane   & 175 & 0.5558 & 0.7425 & 0.7296 & \textbf{0.7900} \\
Bouncing& 175 & 0.5660 & 0.7367 & 0.7337 & \textbf{0.7811} \\
Jump    & 150 & 0.5664 & 0.7078 & 0.7035 & \textbf{0.7590} \\
Samba   & 175 & 0.5255 & 0.7544 & 0.7559 & \textbf{0.7734}  \\ 
Handstand&175 & 0.5384 & 0.6131 & 0.6118 & \textbf{0.6504}  \\
March   & 250 & 0.5224 & 0.6930 & 0.6887 & \textbf{0.7227} \\
Squat   & 250 & 0.5256 & 0.7316 & 0.7304 & \textbf{0.7726} \\
\hline
\end{tabular}
\end{table}


\section{Conclusion}
We have proposed a novel method for human pose and shape estimation using joint points and silhouettes based on SMPL model from four images with different view-points. SMPL model provides better representation about the 3D human body and the prior information including joint points and silhouettes gives strong cues for human body estimation. Our method consists of two steps: joint points based fitting and silhouettes based fitting.The joint points of the images were firstly predicted by deep learning-based human pose estimation. Then, the pose and shape parameters of a SMPL model were estimated by fitting the SMPL model to the joint points of the four images simultaneously.  Furthermore, we identified the corresponding points on the edge of silhouettes and SMPL model to build a novel energy function from 2D and 3D space.\begin{wrapfigure}{r}{0.4\textwidth}
    \centering
	\includegraphics[height=0.51in]{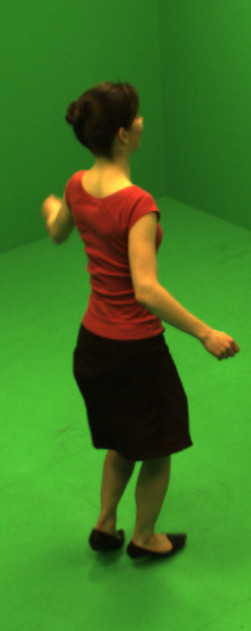}%
	\includegraphics[height=0.51in]{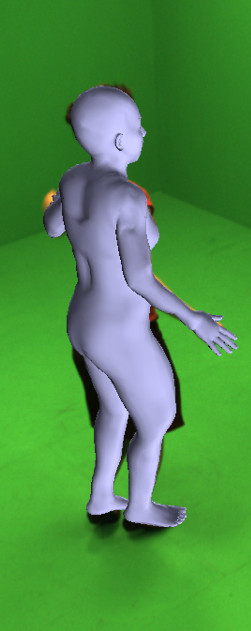}%
	\includegraphics[height=0.51in]{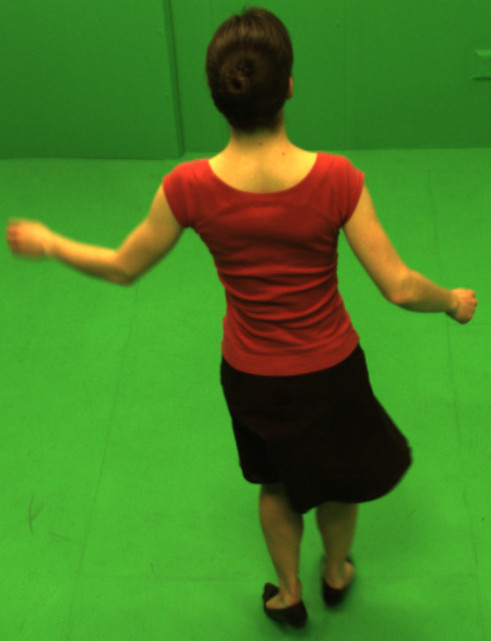}%
	\includegraphics[height=0.51in]{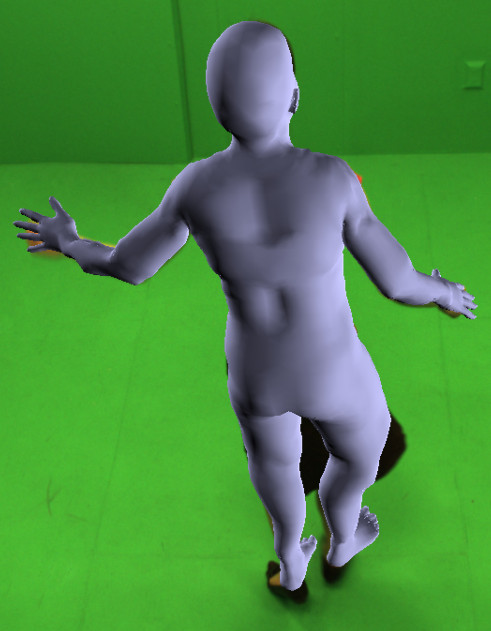}%
	\includegraphics[height=0.51in]{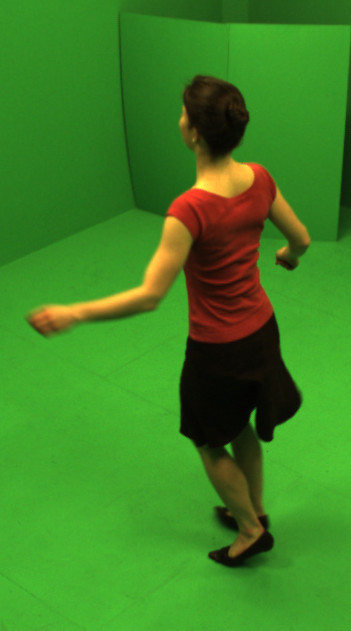}%
	\includegraphics[height=0.51in]{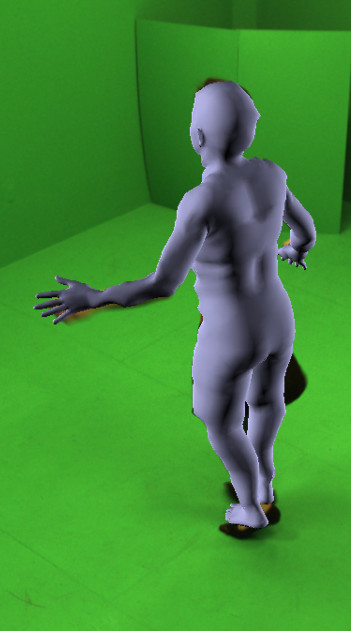}
	
	\includegraphics[height=0.54in]{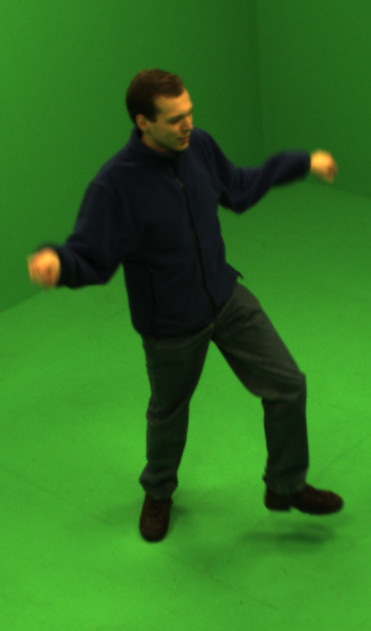}%
	\includegraphics[height=0.54in]{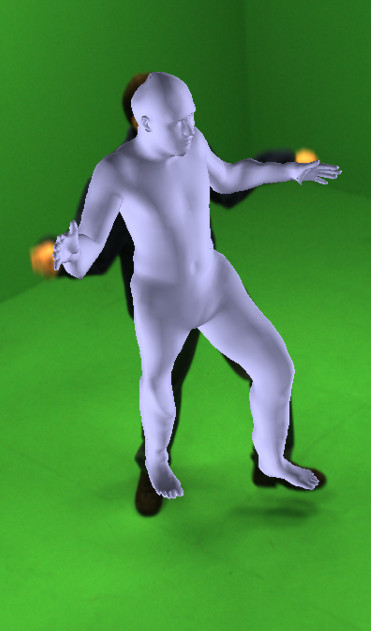}%
	\includegraphics[height=0.54in]{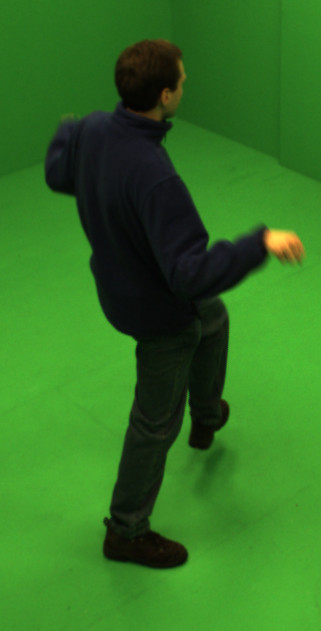}%
	\includegraphics[height=0.54in]{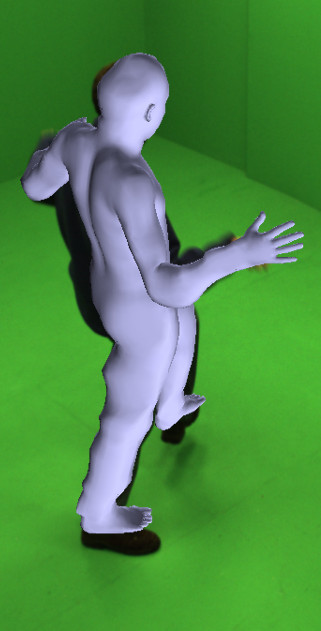}%
	\includegraphics[height=0.54in]{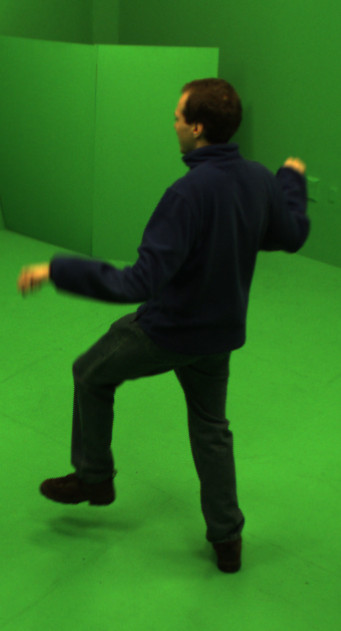}%
	\includegraphics[height=0.54in]{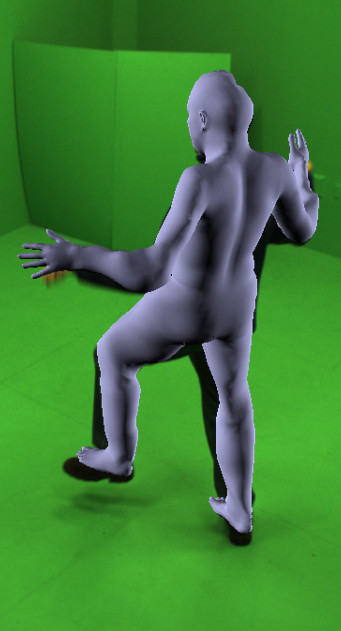}
	
	\includegraphics[height=0.62in]{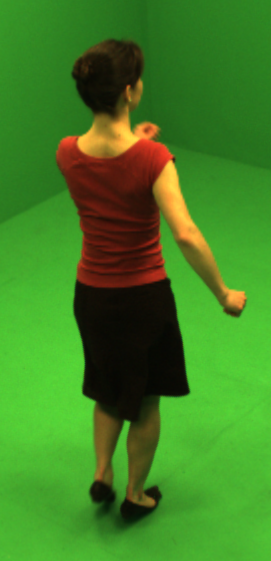}%
	\includegraphics[height=0.62in]{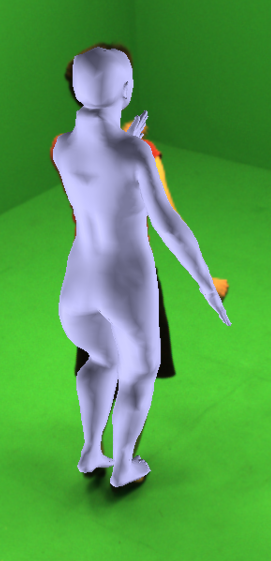}%
	\includegraphics[height=0.62in]{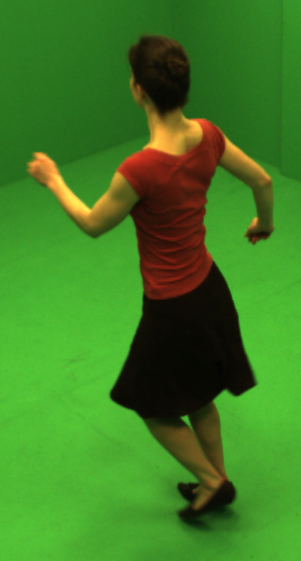}%
	\includegraphics[height=0.62in]{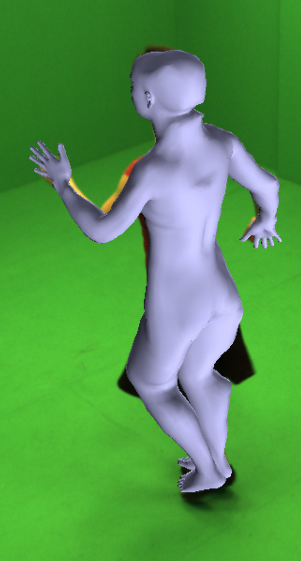}%
	\includegraphics[height=0.62in]{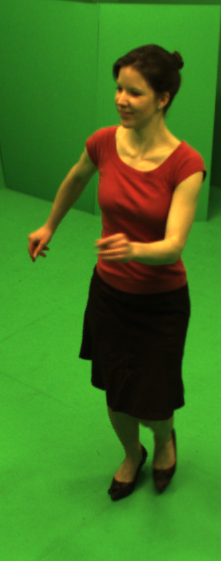}%
	\includegraphics[height=0.62in]{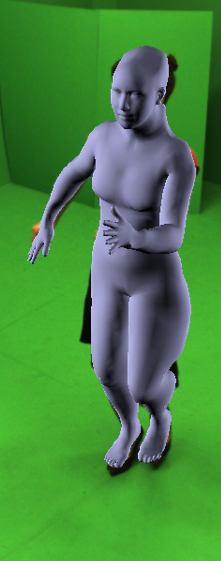}
	
	\includegraphics[height=0.43in]{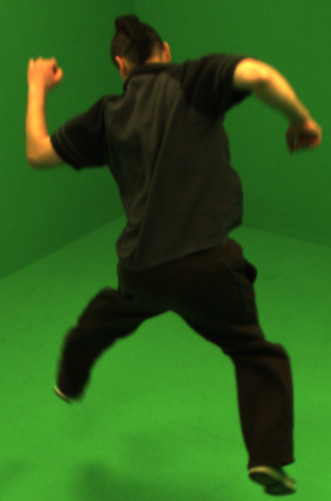}%
	\includegraphics[height=0.43in]{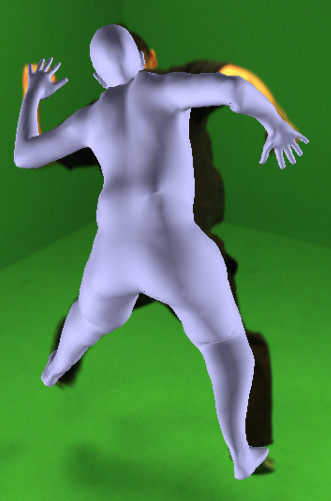}%
	\includegraphics[height=0.43in]{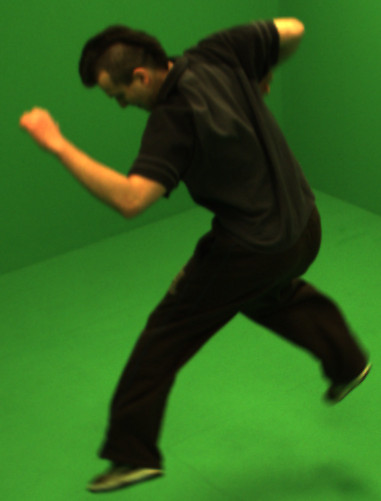}%
	\includegraphics[height=0.43in]{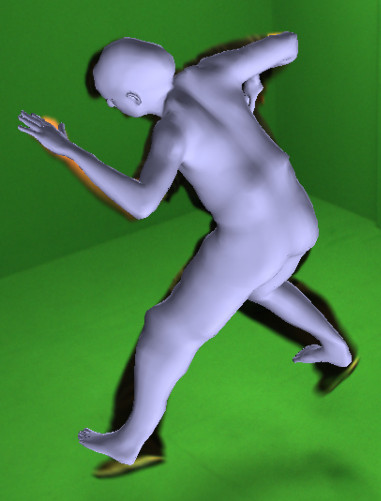}%
	\includegraphics[height=0.43in]{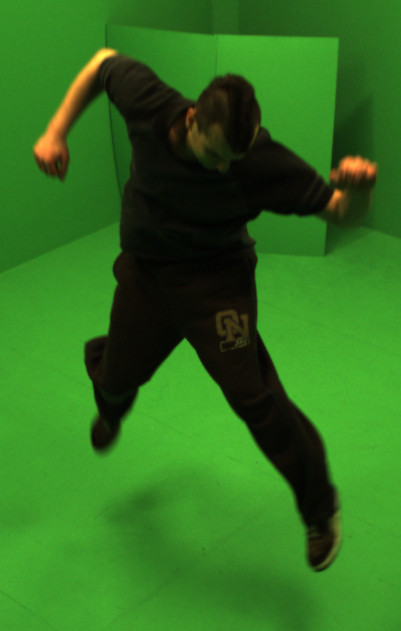}%
	\includegraphics[height=0.43in]{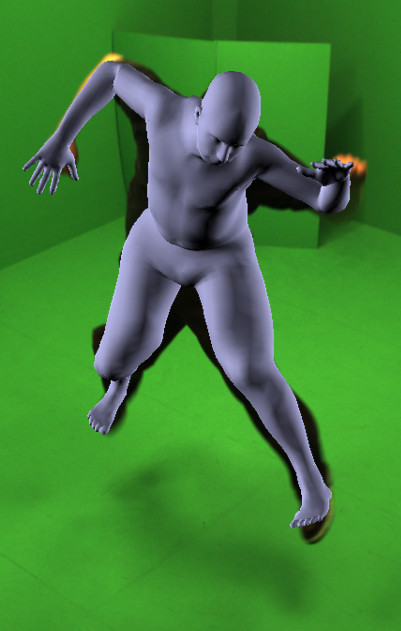}%
	\caption{The results of \textit{Swing, Crane, Samba and Bouncing} from other three views obtained by our method.}
	\label{fig13}
\end{wrapfigure} 
The shape parameters of SMPL were improved by minimizing the novel energy function. Our method not only estimated the pose of the human body, but also obtained better shape appearance of the human body.
The experiments on synthetic and real data indicate that our approach can obtain better human body shape comparing to the previous methods. 
The limitation of our method is that we strongly depend on the estimated joint points and silhouettes, which may result in that the estimation of pose and shape is not correct when the joint points or silhouettes are not predicted correctly. In addition, the texture of the images is not mapped to the 3D model, which makes the appearance not realistic enough in some cases. Overall, our method can be used in many practical fields such as VR video games or biomedical research.  

\bibliographystyle{splncs04}
\bibliography{mybibfile}

\begin{thebibliography}{10}
\providecommand{\url}[1]{\texttt{#1}}
\providecommand{\urlprefix}{URL }
\providecommand{\doi}[1]{https://doi.org/#1}

\bibitem{alldieck_2018}
Alldieck, T., Magnor, M., Xu, W.P., Theobalt, C., Pons-Moll, G.: Video based
  reconstruction of 3d people models. CVPR pp. 8387--8397 (2018)

\bibitem{Anguelov_2005}
Anguelov, D., Srinivasan, P., Koller, D., Thrun, S., Rodgers, J., Davis, J.:
  Scape: Shape completion and animation of people. ACM Trans. on Graph.
  \textbf{24},  408--416 (2005)

\bibitem{xiao2018simple}
Bin, X., Haiping, W., Yichen, W.: Simple baselines for human pose estimation
  and tracking. ECCV pp. 472--487 (2018)

\bibitem{Bogo_ICCV_2015}
Bogo, F., Black, M.J., Loper, M., Romero, J.: Detailed full-body
  reconstructions of moving people from monocular {RGB-D} sequences. ICCV pp.
  2300--2308 (2015)

\bibitem{Bogo_2016}
Bogo, F., Kanazawa, A., Lassner, C., Gehler, P.V., Romero, J., Black, M.J.:
  Keep it {SMPL:} automatic estimation of 3d human pose and shape from a single
  image. ECCV pp. 561--578 (2016)

\bibitem{Dou_2016}
Dou, M.S., Khamis, S., Degtyarev, Y., Davidson, P., Ryan, S.F., Kowdle, A.,
  Escolano, S.O., Rhemann, C., Kim, D., Taylor, J., Vladimir, P.K., Izadi,
  T.S.: Fusion4d: real-time performance capture of challenging scenes. ACM
  Trans. on Graph.  \textbf{35},  114:1--114:13 (2016)

\bibitem{Geman_1987}
Geman, S., McClure, D.: Statistical methods for tomographic image
  reconstruction. Bulletin of the International Statistical Institute
  \textbf{52},  5--21 (1987)

\bibitem{Guan_2009}
Guan, P., {Weiss}, A., {Bãlan}, A.O., {Black}, M.J.: Estimating human shape
  and pose from a single image. ICCV pp. 1381--1388 (2009)

\bibitem{Huang_2017}
Huang, Y., Bogo, F., Lassner, C., Kanazawa, A., Gehler, P.V., Romero, J.,
  Akhter, I., Black, M.J.: Towards accurate marker-less human shape and pose
  estimation over time. 2017 International Conference on 3D Vision (3DV) pp.
  421--430 (2017)

\bibitem{Innmann_2016}
Innmann, M., Zollhöfer, M., Nießner, M., Theobald, C., Stamminger, M.:
  {VolumeDeform}: {Real}-{Time} {Volumetric} {Non}-rigid {Reconstruction}. ECCV
  pp. 362--379 (2016)

\bibitem{Izadi2011}
Izadi, S., Kim, D., Hilliges, O., Molyneaux, D., Newcombe, R., Kohli, P.,
  Shotton, J., Hodges, S., Freeman, D., Davison, A., Fitzgibbon, A.:
  Kinectfusion: Real-time 3d reconstruction and interaction using a moving
  depth camera. Proceedings of the 24th annual ACM symposium on User interface
  software and technology(UIST) pp. 559--568 (2011)

\bibitem{hmrKanazawa17}
Kanazawa, A., Black, M.J., Jacobs, D.W., Malik, J.: End-to-end recovery of
  human shape and pose. CVPR pp. 7122--7131 (2018)

\bibitem{kolotouros_2019}
Kolotouros, N., Pavlakos, G., Black, M.J., Daniilidis, K.: Learning to
  reconstruct 3d human pose and shape via model-fitting in the loop. ICCV pp.
  2252--2261 (2019)

\bibitem{Law2018}
Law, H., Deng, J.: Cornernet: Detecting objects as paired keypoints.
  International Journal of Computer Vision pp. 1--15 (2019)

\bibitem{Li_2019}
Li, Z., Heyden, A., Oskarsson, M.: Parametric model-based 3d human shape and
  pose estimation from multiple views. 21st Scandinavian Conference on Image
  Analysis (SCIA) pp. 336--347 (2019)

\bibitem{Loper_2015}
Loper, M., Mahmood, N., Romero, J., Pons-Moll, G., Black, M.J.: Smpl: A skinned
  multi-person linear model. ACM Trans. on Graph.  \textbf{34},  248:1--248:16
  (2015)

\bibitem{Loper_2014}
Loper, M.M., Black, M.J.: {OpenDR}: An approximate differentiable renderer.
  ECCV pp. 154--169 (2014)

\bibitem{Newcombe2015}
Newcombe, R.A., Fox, D., Seitz, S.M.: Dynamicfusion: Reconstruction and
  tracking of non-rigid scenes in real-time. CVPR pp. 343--352 (2015)

\bibitem{Pavlakos_2019}
Pavlakos, G., Choutas, V., Ghorbani, N., Bolkart, T., Osman, A., Tzionas, D.,
  Black, M.J.: Expressive body capture: 3d hands, face, and body from a single
  image. CVPR pp. 10975--10985 (2019)

\bibitem{pavlakos2018}
Pavlakos, G., Zhu, L.Y., Zhou, X.W., Daniilidis, K.: Learning to estimate 3{D}
  human pose and shape from a single color image. CVPR pp. 459--468 (2018)

\bibitem{pavlakos2019}
Pavlakos, G., Kolotouros, N., Daniilidis, K.: Texturepose: Supervising human
  mesh estimation with texture consistency. ICCV pp. 803--812 (2019)

\bibitem{LeonidNIPS2007}
Sigal, L., Balan, A., Black, M.J.: Combined discriminative and generative
  articulated pose and non-rigid shape estimation. NIPS pp. 1337--1344 (2008)

\bibitem{Slavcheva2017}
Slavcheva, M., Baust, M., Cremers, D., Ilic, S.: Killingfusion: Non-rigid 3d
  reconstruction without correspondences. CVPR pp. 5474--5483 (2017)

\bibitem{varol18_2018}
Varol, G., Ceylan, D., Russell, B., Yang, J., Yumer, E., Laptev, I., Schmid,
  C.: Bodynet: Volumetric inference of {3D} human body shapes. ECCV pp. 20--38
  (2018)

\bibitem{Vlasic2008}
Vlasic, D., Baran, I., Matusik, W., Popovi\'{c}, J.: Articulated mesh animation
  from multi-view silhouettes. ACM Trans. on Graph.  \textbf{27},  97:1--97:9
  (2008)

\bibitem{Weiss_2011}
{Weiss}, A., {Hirshberg}, D., {Black}, M.J.: Home 3d body scans from noisy
  image and range data. ICCV pp. 1951--1958 (2011)

\bibitem{Xu2019}
{Xu}, L., {Su}, Z., {Han}, L., {Yu}, T., {Liu}, Y., {FANG}, L.:
  Unstructuredfusion: Realtime 4d geometry and texture reconstruction using
  commercial rgbd cameras. IEEE Trans. on Pattern Analysis and Machine
  Intelligence pp.~1--1 (2019)

\bibitem{Weipeng_2017}
Xu, W.P., Chatterjee, A., Zollh{\"{o}}fer, M., Rhodin, H., Mehta, D., Seidel,
  H., Theobalt, C.: Monoperfcap: Human performance capture from monocular
  video. ACM Trans. on Graph.  \textbf{37},  27:1--27:15 (2016)

\bibitem{Ye_2012_ECCV}
Ye, G.Z., Liu, Y.B., Hasler, N., Ji, X.Y., Dai, Q.H., Theobalt, C.: Performance
  capture of interacting characters with handheld kinects. ECCV pp. 828--841
  (2012)

\bibitem{Yu_ICCV_2017}
Yu, T., Guo, K.W., Xu, F., Dong, Y., Su, Z.Q., Zhao, J.H., Li, J.G., Dai, Q.H.,
  Liu, Y.B.: Bodyfusion: Real-time capture of human motion and surface geometry
  using a single depth camera. ICCV pp. 910--919 (2017)

\bibitem{Yu_2018_CVPR}
Yu, T., Zheng, Z.R., Guo, K.W., Zhao, J.H., Dai, Q.H., Li, H., Pons-Moll, G.,
  Liu, Y.B.: Doublefusion: Real-time capture of human performances with inner
  body shapes from a single depth sensor. CVPR pp. 7287--7296 (2018)

\end{thebibliography}
%




\end{document}